\documentclass[12 pt]{article}

\pdfoutput=1
\usepackage{amsthm,amsmath,amssymb}
\usepackage[margin=0.75in]{geometry}
\RequirePackage{hyperref}
\usepackage[utf8]{inputenc} %unicode support
\usepackage{graphicx}
\usepackage{algorithmicx}
\usepackage{algpseudocode}
\usepackage{algorithm}
\usepackage{multirow}
\usepackage{nameref}
\usepackage{float}
\usepackage{xcolor}
\usepackage{authblk}

\begin{document}
\title{\sc optimalFlow: Optimal-transport approach to flow cytometry gating and population matching}

\author[1]{Eustasio del Barrio}
\author[2]{Hristo Inouzhe}
\author[3]{Jean-Michel Loubes}
\author[4]{Carlos Matrán}
\author[5]{Agustín Mayo-Íscar}
\affil[1,2,4,5]{\it Departamento de Estadística e Investigación Operativa and IMUVA,
Universidad de Valladolid, Spain.}
\affil[3]{\it Université de Toulouse, Institut de Mathématiques de Toulouse, France.}

\date{}

\date{}
\maketitle

\maketitle

\begin{abstract} % abstract
Data obtained from Flow Cytometry present pronounced variability due to biological and technical reasons. Biological variability is a well-known phenomenon produced by measurements on different individuals, with different characteristics such as illness, age, sex, etc. The use of different settings for measurement, the variation of the conditions during experiments and the different types of flow cytometers are some of the technical causes of variability. This mixture of sources of variability makes the use of supervised machine learning for identification of cell populations difficult. The present work is conceived as a combination of strategies to facilitate the task of supervised gating.

We propose \textit{optimalFlowTemplates}, based on a \textit{similarity distance} and \textit{Wasserstein barycenters}, which clusters cytometries and produces prototype cytometries for the different groups. We show that supervised learning, restricted to the new groups, performs better than the same techniques applied to the whole collection. We also present \textit{optimalFlowClassification}, which uses a database of gated cytometries and optimalFlowTemplates to assign cell types to a new cytometry. We show that this procedure can outperform state of the art techniques in the proposed datasets. Our code is freely available as \emph{optimalFlow} a Bioconductor R package at \href{https://bioconductor.org/packages/optimalFlow}{https://bioconductor.org/packages/optimalFlow}.

optimalFlowTemplates+optimalFlowClassification addresses the problem of using supervised learning while accounting for biological and technical variability. Our methodology provides a robust automated gating workflow that handles the intrinsic variability of flow cytometry data well. Our main innovation is the methodology itself and the optimal-transport techniques that we apply to flow cytometry analysis.
\end{abstract}

%\begin{keyword}
%\kwd{flow cytometry gating}
%\kwd{optimal-transport}
%\kwd{Wasserstein distance}
%\kwd{clustering}
%\kwd{supervised classification}
%\end{keyword}

\section*{Background}
Flow cytometry (FC) works with `high-dimensional quantitative measurement of light scatter and fluorescence emission properties of hundreds of thousands of individual cells in each analysed sample' (see \cite{rev_citos_1}). These quantitative measurements allow to analyse and classify individual cells, facilitating diverse applications. For example, as mentioned in \cite{rev_citos_2}, `flow cytometry is used to identify and
quantify populations of immune cells' in order to monitor the immune state of patients or to detect relevant biomarkers by comparing flow cytometries from different patient groups.

A main component in FC is gating, the assignment of individual cells (data records) into discrete cell types. Manual gating, i.e.,  an expert assigning cell types (labels) to individual cells, using a set of rules on one or two-dimensional projections, has been the prevalent option. However, this manual approach has some shortcomings. Firstly, it is subjective since it depends on the expertise of the user, on the sequence of markers (measured variables) used to do the projections and on the locations of the gates on those projections. Secondly, it can be very time consuming because it is `roughly quadratic in the number of markers' (see \cite{Li2017}). Lastly, the recent increase in the number of markers and number of cells per cytometry makes human error a relevant factor.

In order to avoid some of the difficulties related to manual gating there have been different approaches to automated gating. Some are unsupervised; therefore, there is no use of previously gated cytometries. Hence, gating is done through a clustering procedure. We present a small selection of these unsupervised automated gating procedures. CCST \cite{CCST}, which uses a nonparametric mixture model clustering and a data-derived decision tree representation for gating; FLOCK \cite{FLOCK}, which does grid-based density estimation (with merging) and then k-means; FLAME \cite{FLAME}, which performs skew $t$ model-based clustering and flowClust \cite{flowClust_1,flowClust_2}, which does robust based clustering through $t$ mixture models with Box-Cox transformation. Other related clustering procedures are: flowPeaks \cite{flowPeaks} which performs Gaussian mixture model based clustering (with modified covariances) and merging and flowMeans \cite{flowMeans} which does k-means with initialization via mode detection through kernel density based estimation. More information about state of the art methods can be found in \cite{rev_citos_1, rev_citos_2}.

Accuracy of cell type assignation can be improved using supervised machine learning where historical information is contained in previously gated cytometries (manually or otherwise). Recently, some methods have been produced addressing this problem. In \cite{Li2017}, DeepCyTOF was introduced, essentially combining de-noising, deep-learning algorithms and domain adaptation. In \cite{flowLearn}, flowLearn was introduced, combining density features of the data, manually selected gating thresholds and derivative-based density alignments. We stress that other more classical approaches for supervised learning are also available. For example, random forest algorithms, support vector machines or quadratic discriminant analysis can be used when learning from some previously gated cytometry. Supervised machine learning is a well-documented topic and for more detailed explanations we refer to \cite{alpaydin2014}.

There are two main setups for using supervised learning in the FC context which are relevant in practical studies. Firstly, the classical one, where there is an available data base of historical information. This means that a collection of gated flow cytometries is available and we want to use this information in order to gate a new cytometry. In a second scenario, we have a collection of ungated cytometries and we want to gate manually (or otherwise) a minimal amount of them and use these gated cytometries to classify the rest of the cytometries in the collection. In both setups, there is a fundamental problem intrinsic to FC. That is, flow cytometry data have considerable technical and biological variability which makes the use of supervised learning difficult. Biological variability is due to intrinsic differences between individuals such as health status, age, gender, etc. Technical variability appears through the use of different experimental adjustments, variation of conditions during experiments and the use of different measuring devices (flow cytometers).
%
%Our methodology is the following: as the intrinsic variability of flow cytometry data is high it should be apparent in a database of gated cytometries. Therefore, it is desirable to divide the database, producing groups of cytometries which are more homogeneous. As the cytometries in a given group are more similar to each other, a consensus of these cytometries, an artificial (a prototype, a template) cytometry, should be a good representative to learn from. Again, given the natural variability of flow cytometry, a new cytometry might not resemble some of the prototypes, so it is natural to assign this cytometry to the most similar prototype or group of most similar cytometries and only use these for later supervised learning. In the case that the new cytometry does not resemble any of the prototypes it would not be advisable to use supervised learning. In that case exploratory techniques would be more recommendable.

In this work we provide novel methods for grouping (clustering) gated cytometries. The goal is to produce groups (clusters) of cytometries that have lower variability than the whole collection and, furthermore, that are coherent enough to be considered as a typology by themselves. This, in turn, allows to improve greatly the performance of any supervised learning procedure. We provide evidence of this below. Once we have a partition (clustering) of a collection of cytometries, we provide several methods for obtaining an artificial cytometry (prototype, template) that represents in some optimal way the cytometries in each respective group. These prototypes can be used, among other things, for matching populations between different cytometries as suggested in \cite{flowMatch, flowMap-FR} or for analysing and extracting characteristics of a group of similar cytometries. In addition, a procedure able to group similar cytometries could help to detect individuals with a common particular condition, i.e, a particular sickness such as cancer. In our work we show that this indeed happens.

\textit{optimalFlowTemplates} is our procedure for clustering cytometries and obtaining templates. It is based on recent developments in the field of optimal transport such as a \textit{similarity distance} between clusterings, introduced in \cite{comp_clust_wasser}, and a \textit{barycenter} (Frechet mean, see \cite{bary_wasser, boissard2015}) and  \textit{k-barycenters} (see \cite{bary_calc,bary_wassesr_rob,bary_consensus}) of probability distributions.

Additionally, we introduce a supervised classification tool, \textit{optimalFlowClassification}, for the case when a database of gated cytometries is available. The procedure uses the prototypes obtained by optimalFlowTemplates on the database. These are used to initialise \textit{tclust}, a robust extension of k-means that allows for non-spherical shapes, to gate a new cytometry (see \cite{tclust}, not to be confused with TCLUST, \cite{dost2011}). By using a similarity distance between the best clustering obtained by \textit{tclust} and the artificial cytometries provided by   optimalFlowTemplates we can assign the new cytometry to the most similar template (and the respective group of cytometries). We provide several options of how to assign cell types to the new cytometry using the most relevant information, represented by the assigned template and the respective cluster of cytometries.

\section*{Methods}
We can view a gated flow cytometry, $X^i$, as a collection of $n_i$ multidimensional points with 
their associated labels (cell types or group labels) forming a set $L^i=\{L_k^i\}_{k=1}^{k_i}$ of $k_i$ different labels. Hence, a gated cytometry can be described as $X^i = \{(X^i_j,Y^i_j)\}_{j = 1}^{n_i}$ where $X^i_j\in\mathbb{R}^d$ and $Y^i_j\in L^i$. Alternatively, we could describe it as a 
partition (clustering) of all $X^i_j$ into groups (clusters) formed by points sharing the same labels. That is, $\mathcal{C}^i=\{(\mathcal{C}^i_k, p^i_k)\}_{k=1}^{k_i}$ where $\mathcal{C}^i_k = \{X^i_j:1\leq j\leq n_i,Y_j^i = L^i_k\}$ is a cluster and $p^i_k$ is a weight associated with label $L_k^i$. A third useful description is to view a gated cytometry as a clustering but coming from a mixture of location-scatter multivariate distributions. With some abuse of notation $\mathcal{C}^i = \{(m^i_k,S^i_k,p^i_k)\}_{k=1}^{k_i}$ where $m_k^i,S^i_k$ are the multivariate mean and covariance of the points in cluster $\mathcal{C}^i_k$.
\begin{figure*}[h!]
\begin{center}
\includegraphics[scale = 0.19]{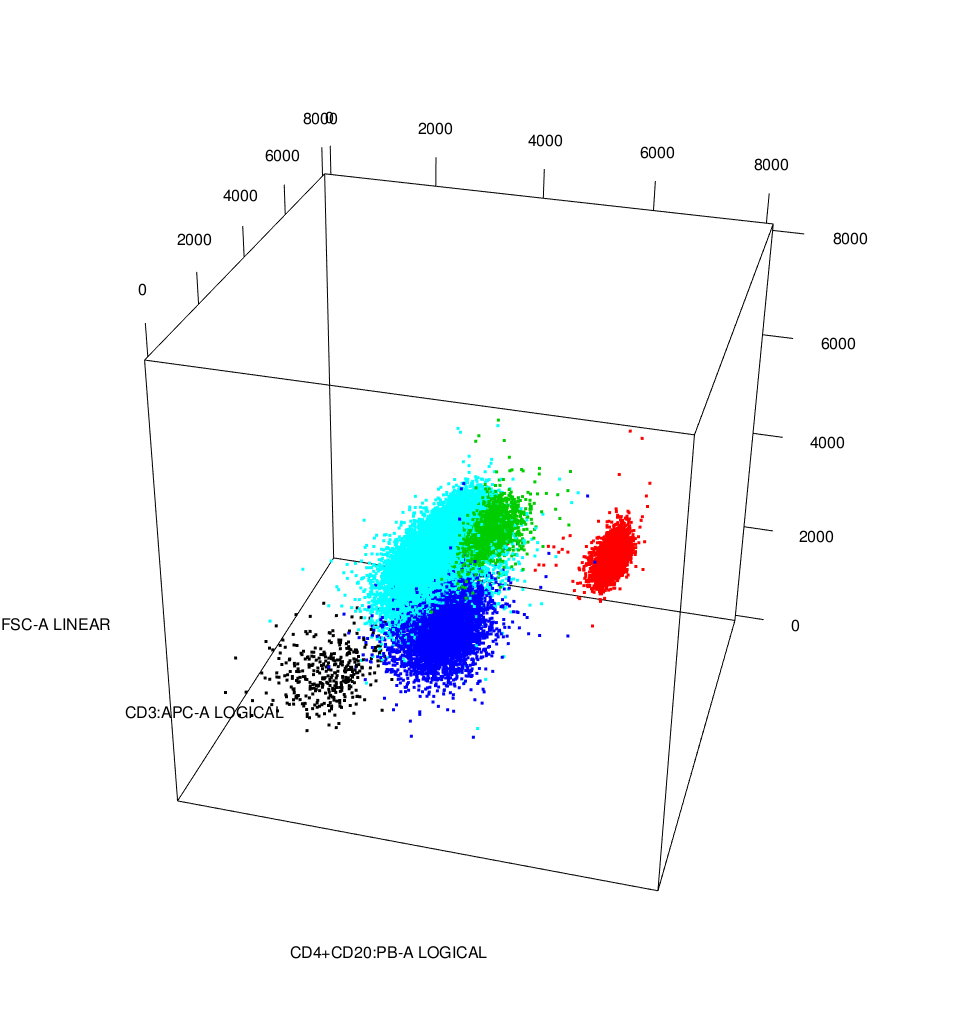}\includegraphics[scale = 0.19]{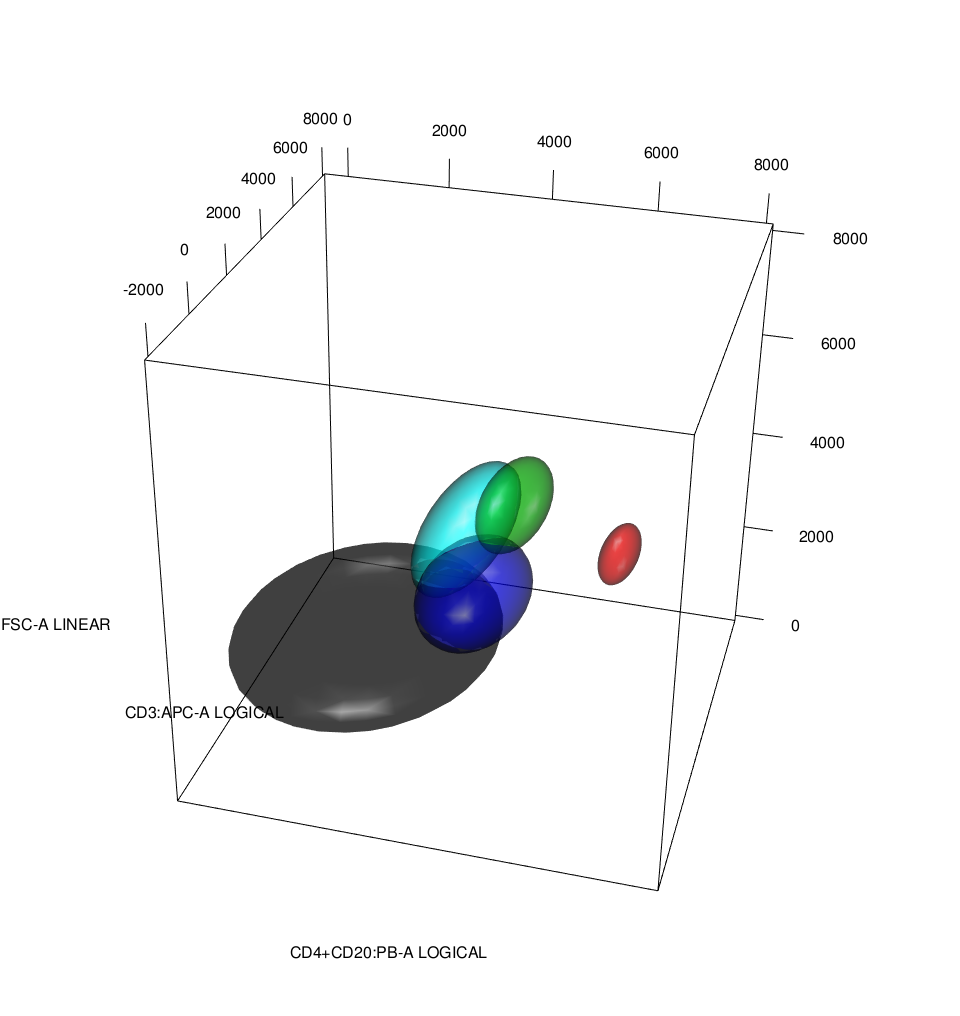}
\end{center}
\caption{A flow cytometry with five cell types viewed in a three-dimensional projection: left as points with labels and right as ellipsoids containing 95$\%$ probability of multivariate normal distributions. We have Basophils in black, CD4+CD8-in red, Eosinophils in green, Monocytes in blue and Neutrophils in cyan.}\label{plot_cyto_view}
\end{figure*}

We provide an example of the different descriptions in Figure \ref{plot_cyto_view}. We have five cell types, hence $L^1=\{Basophils\,(black),\,CD4+CD8-\,(red),\,Eosinophils\allowbreak \,(green),\allowbreak Monocytes\,(blue),\allowbreak\,Neutrophils\,\allowbreak(Cyan)\}$. We have a three-dimensional projection onto three different markers. We can interpret the image on the left as a plot of the coordinates of every cell with its label, but also as the plot of the group of cells labelled as Basophils (black group), and so on... On the other hand, the plot on the right is a representation of the ellipsoid containing $95\%$ of the probability when we see each cluster as a multivariate normal distribution with mean and covariance corresponding to the empirical mean and covariance. As we see from the plots, all of the above descriptions seem to represent the data at hand well and, therefore, all of them could be useful for different applications.
\subsection*{Obtaining prototypic cytometries: optimalFlowTemplates}
Due to the the high variability in flow cytometry data we should expect that learning from different elements in the database should produce significantly different results on the classification of a new cytometry $X^T=\{X^T_1,\dots, X^T_{n_T}\}\subset \mathbb{R}^d$. Our approach is to search for clusters of existing cytometries in the database. In this way we pursue a notable reduction of variability, thus allowing a good representation of the cytometries in each of these groups through prototypic cytometries. Therefore, using a prototype of a group for learning should produce a similar result for classifying $X^T$ to the one obtained using any other cytometry in the same group.
\subsubsection*{Clustering cytometries}
Since gated cytometries can be viewed as partitions (clusterings) and we want to cluster cytometries in order to reduce variability, we want to do clustering of clusterings, also known as metaclustering. The methodology we will develop consists on using some meaningful distance between partitions and then applying hierarchical clustering methods. We use hierarchical clustering since it does not rely on a particular distance and therefore it is well suited for handling a variety of distances between objects. This is not the case in many other usual clustering procedures. 

As a distance between clusterings we propose to use the \textit{similarity distance} (\ref{sim_distance}), introduced in \cite{comp_clust_wasser}. It is based on two auxiliary distances.  The optimal transport distance between two partitions $\mathcal{C}^i$ and $\mathcal{C}^j$, defined as
\begin{equation*}
d_{OT}(\mathcal{C}^i,\mathcal{C}^{j})=\sum_{k=1}^{k=k_i}\sum_{l=1}^{l=k_{j}}w^*_{kl}d(	\mathcal{C}^i_k,\mathcal{C}^{j}_{l}),
\end{equation*}
where $d(\mathcal{C}^i_k,\mathcal{C}^{j}_{l})$ is a distance between clusters $\mathcal{C}^i_k$ and $\mathcal{C}^j_l$. $(w^*_{kl})$ are the solutions of the optimal transport linear program 
\begin{footnotesize}
\begin{equation}\label{opt_ptoblem}
\begin{array}{ll@{}ll}
\text{minimize} & \sum_{k=1}^{k=k_i}\sum_{l=1}^{l=k_{j}}w_{kl}d(\mathcal{C}^i_k,\mathcal{C}^{j}_{l})&\\
\text{subject to} & w_{kl}\geq 0,  &1\leq k\leq k_i,1\leq l\leq k_j&\\ 
				  & \sum_{l = 1}^{l = k_j}w_{kl} = p^i_k, & 1\leq k\leq k_i&\\
				  & \sum_{k = 1}^{k = k_i}w_{kl} = p^j_l, & 1\leq l\leq k_j&\\
				  & \sum_{k=1}^{k=k_i}\sum_{l=1}^{l=k_{j}}w_{kl} = 1.
\end{array}
\end{equation}
\end{footnotesize}
$d_{OT}$ measures the cost of the optimal way of transforming one partition into the other. For more detailed explanations on optimal transport see \nameref{optimal_transport}. The other auxiliary distance is the naive transport distance, defined as
\begin{equation*}
d_{NT}(\mathcal{C}^i,\mathcal{C}^j) = \sum_{k=1}^{k_i}\sum_{l=1}^{k_j}p^i_kp^j_ld(\mathcal{C}^i_k,\mathcal{C}^j_l).
\end{equation*}
It measures the cost of naively transforming one partition into the other.

The \textit{similarity distance} is defined as the quotient
\begin{equation}
\label{sim_distance}
d_S(\mathcal{C}^i,\mathcal{C}^j) = \frac{d_{OT}(\mathcal{C}^i,\mathcal{C}^j)}{d_{NT}(\mathcal{C}^i,\mathcal{C}^j)}.
\end{equation}
We recall that $0\leq d_S\leq 1$, where $d_S = 0$ means that partitions $\mathcal{C}^i,\mathcal{C}^j$ are represented by the same clusters with the same weights and $d_S = 1$ means that every cluster in $\mathcal{C}^i$ is transported proportionally to every cluster in $\mathcal{C}^j$. Therefore, values of $d_S$ close to 0 can be interpreted as high similarity between clusterings, and values of $d_S$ close to 1 can be interpreted as very dissimilar clusterings.

%Recall that, $d_{OT}$, defined in (\ref{d_ot}),  measures the cost of the optimal way of transforming one partition into the other. On the other hand, the naive transport distance, $d_{NT}$, defined in (\ref{d_nt}), measures the cost of naively transforming one partition into the other.

%The \textit{similarity distance} is defined as the quotient
%\begin{equation}
%%\label{sim_distance}
%d_S(\mathcal{C}^i,\mathcal{C}^j) = \frac{d_{OT}(\mathcal{C}^i,\mathcal{C}^j)}{d_{NT}(\mathcal{C}^i,\mathcal{C}^j)}.
%\end{equation}
%We recall that $0\leq d_S\leq 1$, where $d_S = 0$ means that partitions $\mathcal{C}^i,\mathcal{C}^j$ are represented by the same clusters with the same weights and $d_S = 1$ means that every cluster in $\mathcal{C}^i$ is transported proportionally to every cluster in $\mathcal{C}^j$.

%Therefore, values of $d_S$ close to 0 can be interpreted as high similarity between clusterings, and values of $d_S$ close to 1 can be interpreted as very dissimilar clusterings.

In order to completely define $d_{S}$, we need to specify a distance between clusters. Our choice is to use the well-known Wasserstein distance (see \nameref{optimal_transport}) so 
\begin{equation}\label{wasser_dist_cluster}
d(\mathcal{C}^i_k,\mathcal{C}^{j}_{l}) = \mathcal{W}_2(N(m^i_k,S^i_k), N(m^j_l,S^j_l)).
\end{equation}
In essence, we are treating clusters as multivariate normal distributions, $N(m^i_k,S^i_k)$ and $ N(m^j_l,S^j_l)$, with means and covariances calculated from the clusters. Our choice of the Wasserstein distance is based on the desire to account for the spatial shapes of the clusters and to obtain templates for the groups of cytometries. We stress that all results in this work are also valid when understanding clusters as members of a location-scatter family.

Another interesting measure for cluster difference is, $\mathcal{W}_\gamma(\mathcal{C}^i_k,\mathcal{C}^{j}_{l})$, the (entropy) regularized Wasserstein distance that we have written down in \nameref{optimal_transport} equation (2), where clusters are understood as empirical distributions. We recall that the entropy regularized Wasserstein distance is strictly convex and there are efficient solutions based on the Sinkhorn algorithm (see \cite{cuturi2014}). However, any other dissimilarity measure can be used and several have been used in the context of cluster comparison in FC. For example, the symmetric Kullback-Leibler divergence was used in \cite{flowMatch}, where
\begin{align}\label{dkl}
d_{KL}(\mathcal{C}^i_k,\mathcal{C}^{j}_{l}) &= \frac{1}{2}\Big(KL(N(m^i_k,S^i_k)\| N(m^j_l,S^j_l))\nonumber\\
 &+ KL(N(m^j_l,S^j_l)\| N(m^i_k,S^i_k))\Big),
\end{align}
and the Friedman-Rafsky test statistic was used in \cite{flowMap-FR}.

When we see clusters as collections of points, and we have different clusterings of the same data, the Adjusted Rand Index, the Jaccard distance or other similar can be used, at the expense of losing spatial information.

\begin{algorithm}
\caption{optimalFlowTemplates}
\label{optimalFlowTemplates}
\textbf{Input:} {$X^1,\dots,X^N$, $equal.weights$}
\begin{algorithmic}[1]
\For {$i\leq N$}
	\While{$k\leq k_i$ and $|\mathcal{C}^i_k|$ enough for covariance estimation}
    	\State $m^i_k\gets\mathrm{mean}\,\mathcal{C}^i_k$; $S^i_k \gets \mathrm{cov}\,\mathcal{C}^i_k$
    	\If{$equal.weights = True$}
    		\State $p^i_k \gets 1/k_i$
    	\Else 
    		\State $p^i_k \gets |\mathcal{C}^i_k|/\sum_{k=1}^{k_i}|\mathcal{C}^i_k|$
    	\EndIf
    	\State $\mathcal{C}^i_k\gets (m^i_k,S^i_k,p^i_k)$
    \EndWhile
\EndFor
\For{$i\leq N$}
	\For{$i<j\leq N$}
		\State $S_{ij}\gets d_S(\mathcal{C}^i,\mathcal{C}^j)$
	\EndFor
\EndFor
\State $\mathfrak{T}\gets$ hierarchical clustering with $S$	
\For {$i \leq |\mathfrak{T}|$}
	\State $\mathcal{T}^i\gets$ template obtention on cytometries in $\mathfrak{T}_i$	
\EndFor
\State $\mathcal{T}=\{\mathcal{T}^i,\dots, \mathcal{T}^{|\mathfrak{T}|}\}$
\end{algorithmic}
\textbf{Output:} $\mathfrak{T}, \mathcal{T}$
\end{algorithm}

The clustering of cytometries is presented in lines 1-17 in Algorithm \ref{optimalFlowTemplates}, resulting in a partition, $\mathfrak{T}=\{\mathfrak{T}_1,\dots,\mathfrak{T}_{|\mathfrak{T}|}\}$, of the input cytometries. Lines 12-16 are concerned with the obtention of a distance matrix $S$ that, in line 17, is used to perform hierarchical clustering. Classical agglomerative algorithms can be used, but also density-based algorithms as DBSCAN and HDBSCAN.

\subsubsection*{Template obtention through consensus clustering}\label{Template obtention}
Once we have a partition, $\mathfrak{T}$, of the collection of cytometries $\{\mathcal{C}^j\}_{j=1}^N$, we want to obtain a prototype cytometry, $\mathcal{T}^i$, for every group of cytometries, $i$, in the partition $\mathfrak{T}$ (lines 18-21 in Algorithm \ref{optimalFlowTemplates}). To address this goal, we resort to k-barycenters using Wasserstein distance, which provide a suitable tool for consensus on probability distributions (see \cite{bary_consensus}). We propose three different methods on how to obtain a template cytometry from a group of cytometries, that is, on how to do consensus (ensemble) clustering on flow cytometries. These methods are given in Algorithms \ref{pooling}, \ref{density_based} and \ref{kbar_algor}.

\begin{algorithm}
\caption{Pooling. Only possible when $\{L^i\}_{i=1}^N\subset L = \{L_1,\dots, L_K\}$. This is the case for a set of gated cytometries with identified cell populations.}
\label{pooling}
\textbf{Input:} {$\mathcal{C}^1,\dots,\mathcal{C}^N$, $\mathfrak{T}$}
\begin{algorithmic}[1]
%\For {$i\leq |\mathfrak{T}|$}
	\For {$j\leq K$}
			\State {$C_{ij}\gets$ set of all clusters associated with label $L_j$ for the cytometries in group $\mathfrak{T}_i$.}
			\If{$|C_{ij}|>0$}
				\State{$\mathcal{T}^i_j\gets$ take 1-barycenter of the clusters in $C_{ij}$ viewed as multivariate normals.}
			\Else
				\State $\mathcal{T}^i_j$ is empty
			\EndIf
	\EndFor
	\State $\mathcal{T}^i\gets\{\mathcal{T}^i_1,\dots,\mathcal{T}^i_K\}$
%\EndFor
%\State $\mathcal{T}\gets\{\mathcal{T}^1,\dots,\mathcal{T}^{|\mathfrak{T}|}\}$
\end{algorithmic}
\textbf{Output:} $\mathcal{T}^i$
\end{algorithm}

\begin{figure*}[h!]
\begin{center}
\includegraphics[scale = 0.18]{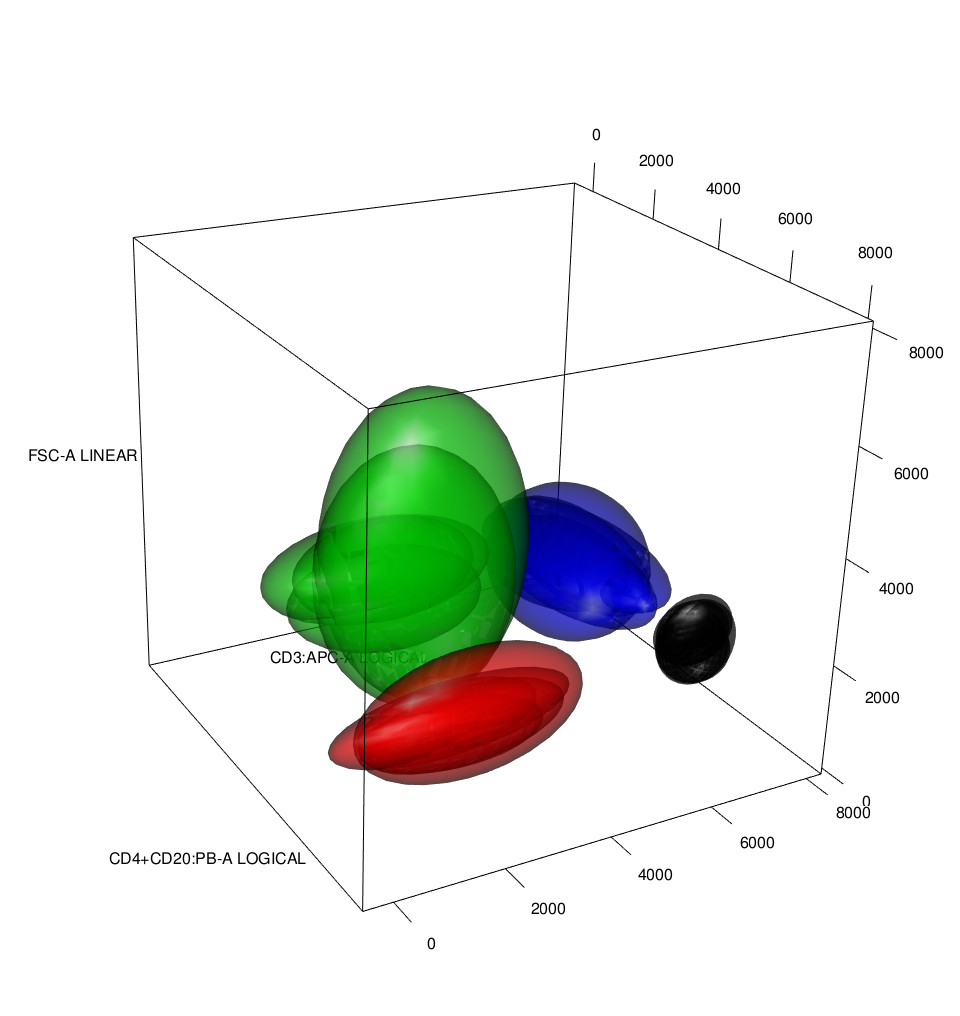}\includegraphics[scale = 0.18]{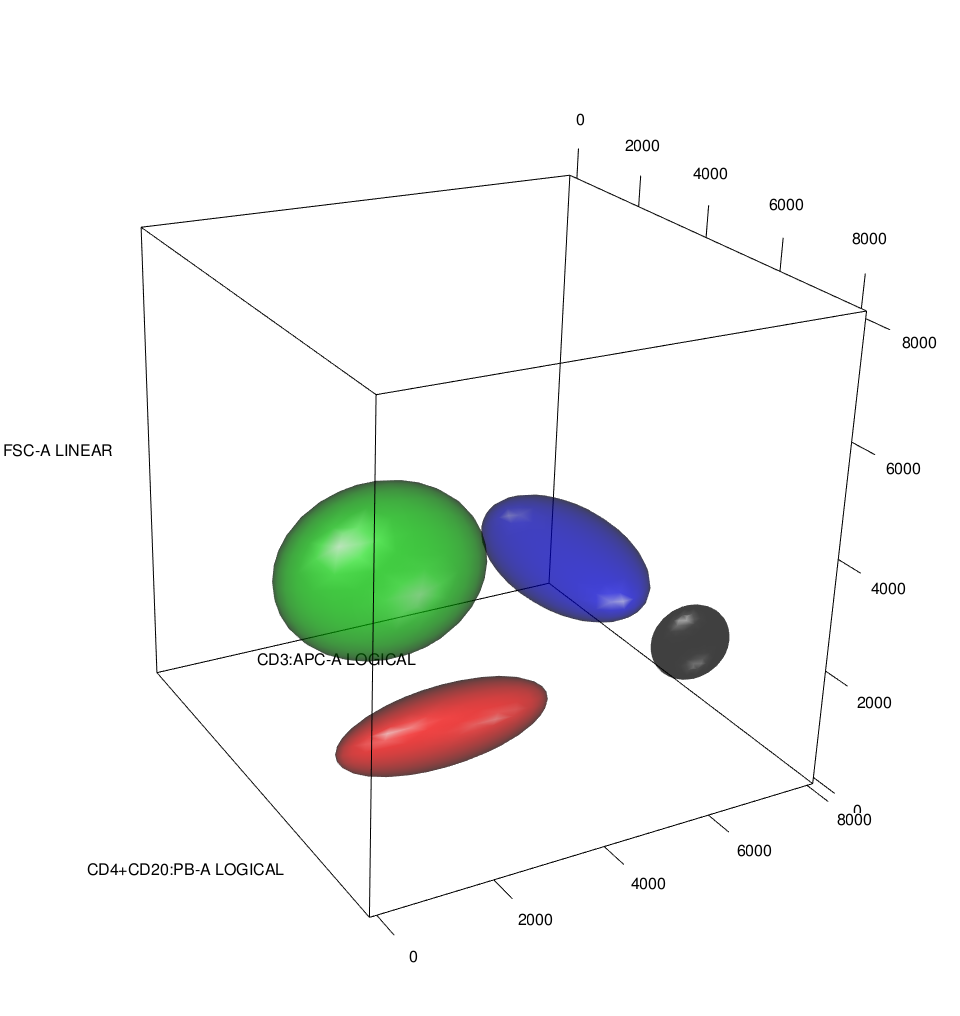}
\end{center}
\caption{An application of Algorithm \ref{pooling}-Pooling. On the left we have 5 different cytometries, each with 4 different identified cell types given by $\{Monocytes\,(black),\, CD4+CD8-\,(red),\, Mature\,Sig\,Kappa\,(green),\,TCRgd-\,(blue)\}$. On the right we have a prototype cytometry obtained taking the 1-barycenter for each cell type. Ellipsoids contain 95\% of the probability of the respective multivariate gaussian distributions.}\label{plot_pooling}
\end{figure*}

\begin{algorithm}
\caption{Density based hierarchical clustering}
\label{density_based}
\textbf{Input:} {$\mathcal{C}^1,\dots,\mathcal{C}^N$, $\mathfrak{T}$}
\begin{algorithmic}[1]
%\For {$i\leq |\mathfrak{T}|$}
	\State	$C^i\gets$ set formed by every cluster of every cytometry in group $\mathfrak{T}_i$.
	\For {$j,k\leq |C^i|$}
		\State $W_{jk} \gets \mathcal{W}_2(N(m^i_j,S^i_j),N(m^i_k,S^i_k))$
	\EndFor
	\State $T\gets$ partition using density based hierarchical clustering on $W$.
	\For{$j\leq |T|$}
		\State $\mathcal{T}^i_j\gets$ barycenter of elements with label $j$ in $T$.
	\EndFor
	\State $\mathcal{T}^i \gets \{\mathcal{T}^i_1,\dots,\mathcal{T}^i_{|T|}\}$
%\EndFor
%\State $\mathcal{T} \gets \{\mathcal{T}^1,\dots,\mathcal{T}^{|\mathfrak{T}|}\}$
\end{algorithmic}
\textbf{Output:} $\mathcal{T}^i$
\end{algorithm}

\begin{figure*}[h!]
\begin{center}
\includegraphics[scale = 0.19]{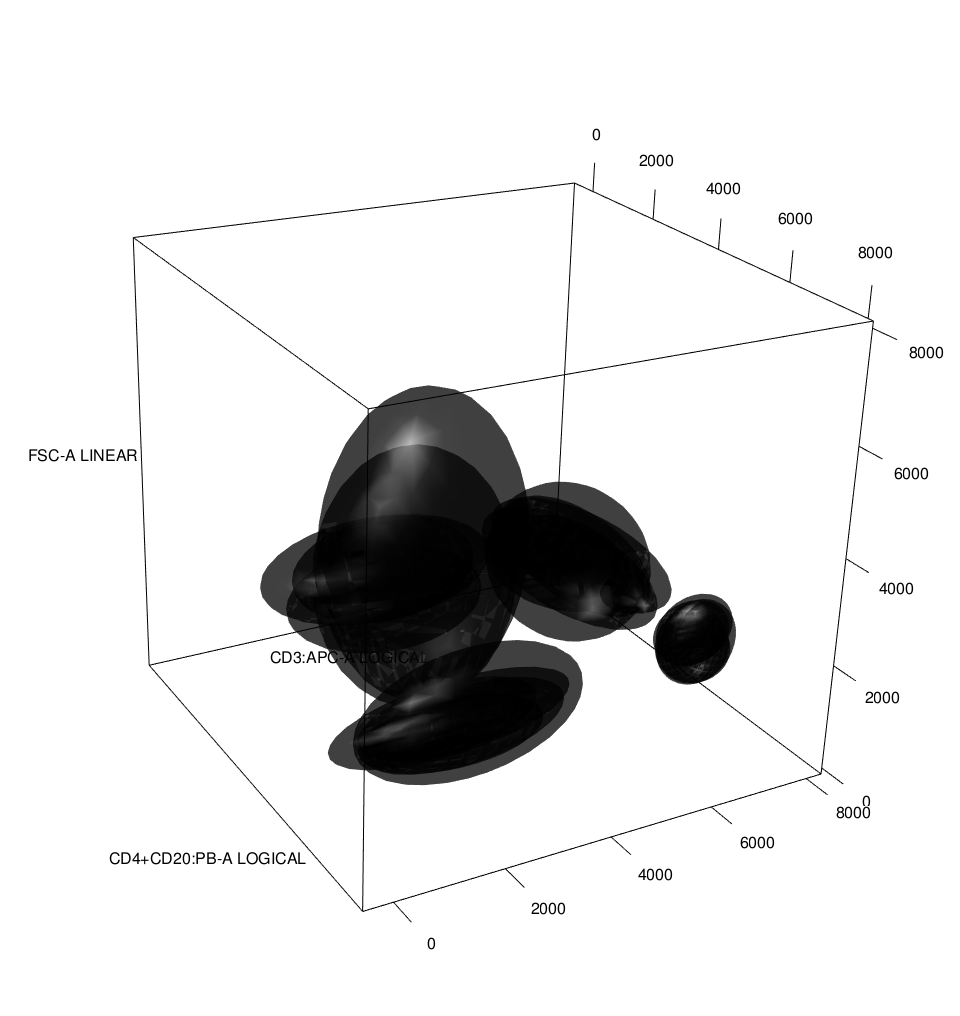}\includegraphics[scale = 0.19]{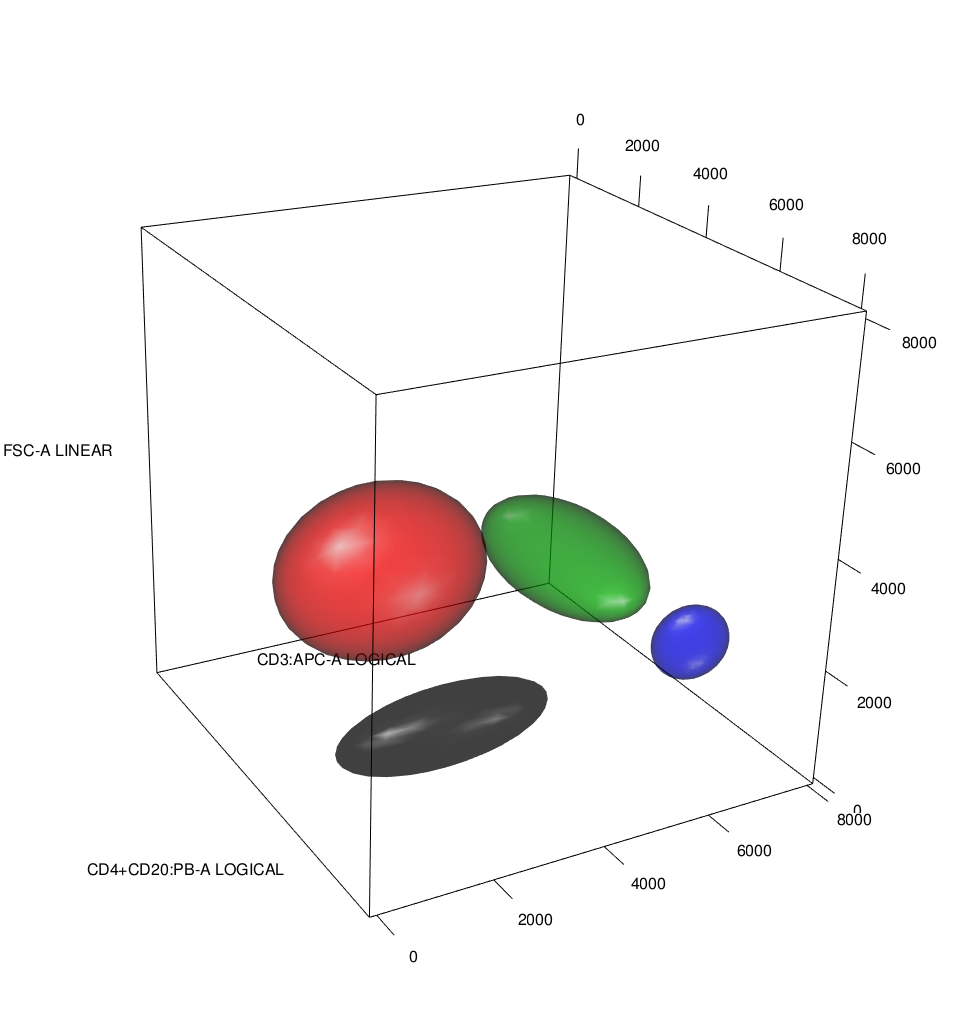}
\end{center}
\caption{Application of Algorithm \ref{density_based} - Density based. On the left we have the same 5 cytometries as in Figure \ref{plot_pooling} but each cytometry is grouped in clusters without cell types being identified. On the right we have a prototype cytometry obtained taking the denisty based hierarchical clustering approach on the cytometries represented on the left. Ellipsoids contain 95\% of the probability of the respective multivariate gaussian distributions.}\label{plot_hdbscan}
\end{figure*}

\begin{algorithm}
\caption{k-barycenter}
\label{kbar_algor}
\textbf{Input:} {$\mathcal{C}^1,\dots,\mathcal{C}^N$, $\mathfrak{T}$, $K$}
\begin{algorithmic}[1]
%\For {$i\leq |\mathfrak{T}|$}
	\State	$C^i\gets$ set formed by every cluster of every cytometry in group $\mathfrak{T}_i$.
	\State $\mathcal{T}^i\gets$ $K$-barycenter of the elements in $C^i$.
%\EndFor
%\State $\mathcal{T} \gets \{\mathcal{T}^1,\dots,\mathcal{T}^{|\mathfrak{T}|}\}$
\end{algorithmic}
\textbf{Output:} $\mathcal{T}^i$
\end{algorithm}

\begin{figure*}[h!]
\begin{center}
\includegraphics[scale = 0.19]{database_normal}\includegraphics[scale = 0.19]{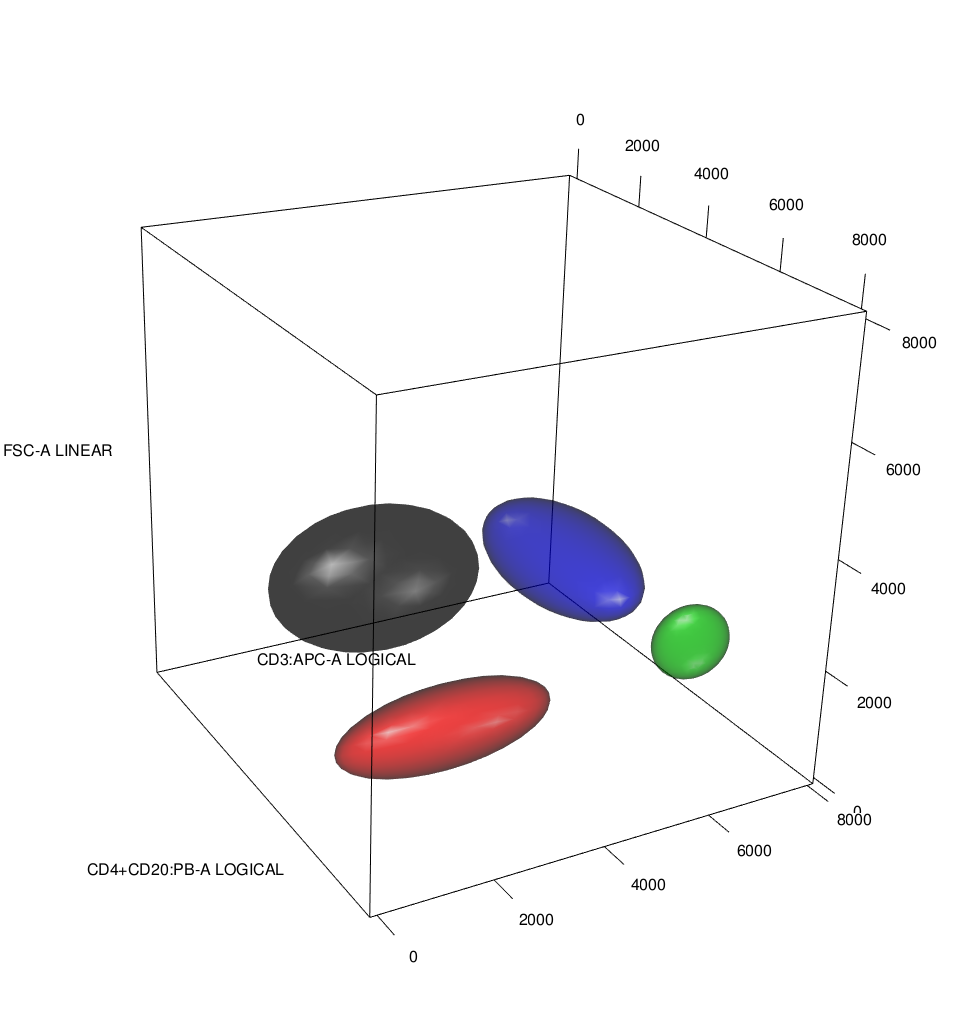}
\end{center}
\caption{Application of Algorithm \ref{kbar_algor} - 4-barycenter. On the left we have the same 5 cytometries as in Figure \ref{plot_pooling} but each cytometry is grouped in clusters without cell types being identified. On the right we have a prototype cytometry obtained taking the 4-barycenter of the cytometries represented on the left. Ellipsoids contain 95\% of the probability of the respective multivariate gaussian distributions.}\label{plot_barycenter}
\end{figure*}

The intention behind pooling (Algorithm \ref{pooling}), is to take advantage of the fact that we have groups of similar cytometries and that cell types are known. A prototype of a cell type is obtained through a (1-)barycenter, a consensus representation, of the multivariate distributions that represent the same cell type in the cytometries that are members of the same group in $\mathfrak{T}$.  A prototype cytometry is the collection of prototypes of each cell type. This can be seen in Figure \ref{plot_pooling}. In the left hand side we have 5 different cytometries where each has 4 different cell types, hence $L = \{Monocytes\,(black),\, CD4+CD8-\,(red),\, Mature\,Sig\,Kappa\,(green),\,TCRgd-\,(blue)\}$. As cell types are known, we take all the black ellipsoids of the left plot, representing the different normal distributions, and obtain the black ellipsoid on the right plot, the barycenter of the group of normal distributions, as a consensus element for Monocytes. Doing this for every cell type gives us the prototype cytometry represented on the right of Figure \ref{plot_pooling}.

However, our templates could be obtained even when we have gated cytometries but without identified cell tpes. This could be the case when unsupervised gating is used to obtain a database of gated cytometries. Density based hierarchical clustering (Algorithm \ref{density_based}) and k-barycenter (Algorithm \ref{kbar_algor}) are based on the idea that clusters that are close in Wasserstein distance should be understood as representing the same, although we may not know which, cell type. When using k-barycenters we have to specify the number of cell types, $K$, that we want for the artificial cytometry. However, when using density based hierarchical clustering as HDBSCAN (see \cite{hdbscan}) or DBSCAN (see \cite{dbscan}) the selection of the number of cell types for the prototype cytometry is automatic. Recall that both k-barycenters, through trimming, and density based hierarchical clustering, are robust clustering procedures.

In Figures \ref{plot_hdbscan} and \ref{plot_barycenter} we have a representation of how Algorithms \ref{density_based} and \ref{kbar_algor} work. Since we do not have cell type information for the 5 gated cytometries, we have the plot that can be seen on the left of Figures \ref{plot_hdbscan} and \ref{plot_barycenter}. However, the absence of this information can be mitigated using the spatial information, which clearly shows a group structure between the ellipsoids. We use density based hierarchical clustering and k-barycenters respectively, to try to capture this spatial information. As a result, we obtain the template cytometries on the right of Figures \ref{plot_hdbscan} and \ref{plot_barycenter}. Clearly, we see that the templates represent well the real cell types behind the cytometries (compare with Figure \ref{plot_pooling}), although we still do not know the cell types corresponding to each ellipsoid. This could be achieved using expert information or matching populations.

\subsection*{Supervised classification: optimalFlowClassification}

Now, our goal is to assign cell types to a new cytometry $X^T$, using the information given in a database of gated cytometries $\{\mathcal{C}^i\}_{i=1}^N$. The different sources of variability, mainly those of a technical nature and those which are properly due to different cohorts present in the database, advise to search for  different cytometric structures. Hence, we should assign $X^T$ to the group of cytometries that is more similar to it and then use supervised techniques. Indeed, this is the purpose of optimalFlowClassification, as shown in Algorithm \ref{optimalFlowClassification}. As an input we apply optimalFlowTemplates to the database $\{\mathcal{C}^i\}_{i=1}^N$ in order to obtain the partition $\mathfrak{T}$ and the templates $\mathcal{T}$.

\begin{algorithm}[h]
\caption{optimalFlowClassification}
\label{optimalFlowClassification}
\textbf{Input:} {$X^T=\{X^T_1,\dots, X^T_{n_T}\}$, $\mathfrak{T}$,$\mathcal{T}$, \emph{initialization}}
\begin{algorithmic}[1]
\If{\emph{initialization} = supervized}
	\For {$i \leq |\mathfrak{T}|$}
		\State $\mathcal{C}^{i,u}\gets$ \textit{tclust} on $X^T$ initialized with $\mathcal{T}^i$	
	\EndFor
	\State $\mathcal{C}^u\gets\arg\max$ of \textit{tclust} objective function over all ${C}^{i,u}$
\Else
	\State $\mathcal{C}^u\gets$ apply some clustering procedure on $X^T$
\EndIf
\For {$i \leq |\mathfrak{T}|$}
	\State $S_i \gets d_S(\mathcal{C}^u,\mathcal{T}^i)$
\EndFor
\State $\mathcal{T}^*\gets \mathcal{T}^{\arg \min S_i}$; $\mathfrak{T}_*\gets \mathcal{T}_{\arg \min S_i}$
\State $\mathcal{C}^T\gets$ labelling of $X^T$ using transfer labelling or supervised classification based on $\mathcal{T}^*$ or $\mathfrak{T}_*$.
\end{algorithmic}
\textbf{Output:} $\mathcal{C}^T$
\end{algorithm}

Lines 1-5 in Algorithm \ref{optimalFlowClassification} are dedicated to finding an unsupervised partition of the new cytometry $X^T$ using as initialization for \textit{tclust} the prototypes of the database. Initializing with the database entries attempts to use the available information optimally. Hence, if $X^T$ is similar to some of the cytometries in the database, appropriate initialization should be advantageous. However, some other suitable unsupervised initializations can be used, such as the ones proposed in FLOCK, flowPeaks or flowMeans. We need to cluster $X^T$ in order to compare it with the template cytometries.

Notice that tclust, \textit{tclust}, introduced in \cite{tclust}, is a more sophisticated version of k-means, allowing ellypsoidal clusters with different sizes and shapes. Similarly to k-means, this robust model-based clustering procedure needs an initialization and its behaviour improves notably if that initialization is well suited. Nonetheless, it is possible to use any other unsupervised procedure that allows an initialization with a clustering defined by probability distributions. For example, this is the case for the popular \textit{mclust} \cite{mclust_1,mclust_2}, a finite Gaussian mixture model-based clustering based on an EM-algorithm. 

\textit{tclust} searches for a partition $\{\mathcal{C}_0,\dots, \mathcal{C}_k\}$ of $X = \{X_1,\dots, X_n\}$, with $|\mathcal{C}_0|=\lceil n\alpha \rceil$, vectors $m_j$, positive definite matrices $S_j$ and weights $p_j\in[0,1]$ that approximately maximize the pseudo-likelihood
\begin{equation}\label{tclust_objective}
\sum_{j=1}^{k}\sum_{i\in \mathcal{C}_j}\log\left( p_j\varphi(X_i;m_j, S_j)\right), 
\end{equation}
under restrictions over the scatter matrices $S_j$. By $\varphi(\cdot;m_j, S_j)$ we denote the density function of the multivariate normal $N(m_j, S_j)$. $\mathcal{C}_0$ is the cluster of trimmed observations, where the trimming level is $\alpha$.

The details of the algorithm can be found in \cite{tclust_algorithm}. For us it is relevant to recall only the initialization step, i.e, to provide an initial $\theta^0=(p^0_1,\dots, p^0_k,m_1^0, \dots, m_k^0, S_1^0,\allowbreak \dots, S_k^0)$. Subsequently, we only need a set of weights with corresponding means and covariances to initialize \textit{tclust}.

We favour the use of \textit{tclust} over k-means since it allows for non-spherical clusters and for trimming, making partitions more robust to outliers and even bridge-points.

However, our procedure is compatible with any other way of obtaining a partition of the data, which we reflect on lines 6-8. In particular, in our experiments we have used flowMeans as an alternative initialization to \emph{tclust}.

In lines 9-12 we look to assign $X^T$, using the clustering $\mathcal{C}^u$, produced in the previous step, to the template that is closest in similarity distance to $\mathcal{C}^u$. With this we hope to use only the most relevant information of the database, summarized in $\mathcal{T}^*$ and $\mathfrak{T}_*$.

The last step in algorithm \ref{optimalFlowClassification}, line 13, is concerned with assigning cell types to $X^T$. To do this we have several options. We can try to relabel $\mathcal{C}^u$ in an optimal way using $\mathcal{T}^*$ or $\mathfrak{T}_{*}$, i.e, do label transfer. Alternatively, we can use $\mathcal{T}^*$ to do Quadratic Discriminant Analysis (QDA) or we can find the most similar partition in similarity distance (\ref{sim_distance}) from $\mathfrak{T}_{*}$ to $\mathcal{C}^u$ and use it to do QDA or random forest classification. In short, we can do label transfer or supervised classification.

For supervised classification we use standard tools, random forest and QDA; however, other methods can be used in a straightforward fashion. We stress that when using QDA and $\mathcal{T}^*$ we are using non-linear multidimensional gating regions obtained from $\mathcal{T}^*$ in order to classify $X^T$. This can be taught as an extension of the method presented in \cite{flowLearn} where only linear one-dimensional regions are used. Another interesting fact is that the use of $d_S$ allows us to select the most similar real cytometry to $C^u$, resulting in supervised tools being more effective.

The problem of relabelling a clustering $\mathcal{C}^j$ with respect to another clustering $\mathcal{C}^i$ is usually stated as a weighted bipartite matching problem, where weights are related to the similarity between clusters in the two partitions. This problem can be solved by the Hungarian method \cite{bip_match}. Generalized edge cover  is another possible solution to relabelling (see \cite{flowMatch}).

Additionally, we introduce an approach to obtain a fuzzy relabelling based on solving the optimal transport linear program associated to (\ref{opt_ptoblem}). The solution, $(w_{kl}^*)$, is the base for this fuzzy relabelling. We define the score of cluster $l$ in $\mathcal{C}^j$ to come from cluster $k$ in $\mathcal{C}^i$ as $s^l_k=w^*_{kl}/p^j_l$. In words, $s^l_k$ is the proportion of probability coming from cluster $k$, with respect to the probability in cluster $l$, that arrives at cluster $l$. Clearly, $0\leq s^l_k\leq 1$, and the closer to 1 the score is the more evidence we have that cluster $k$ and $l$ represent the same cluster. A fuzzy relabelling for cluster $l$ in $\mathcal{C}^j$ is the collection of all the scores $s^l = \{s^l_1,\dots,s^l_{|\mathcal{C}^i|}\}$. A variation of the previous score is $\tilde{s}^l_k = s^l_k*w^*_{kl}/p^i_k$, where we are weighting by the proportion of cluster $k$ that goes to cluster $l$, with respect to the probability contained in cluster $k$. In this way we down-weight the effect of a small proportion of a big cluster with respect to a big proportion of a small cluster arriving to $l$. From these fuzzy relabellings a hard relabelling can be easily obtained.

Again, a suitable distance between clusters can be the Wasserstein distance as in (\ref{wasser_dist_cluster}). However, another possibility is to use 
\begin{equation} \label{wasser_dist_cluster_2}
d(\mathcal{C}^i_k,\mathcal{C}^j_l) = \frac{1}{|\mathcal{C}^i_k||\mathcal{C}^j_l|}\sum_{x\in \mathcal{C}^i_k}\sum_{y\in \mathcal{C}^j_l}\|x-y\|^2.
\end{equation}
(\ref{wasser_dist_cluster}) is computationally very efficient but does not allow for the labelling of very small clusters in $\mathcal{C}^j$.(\ref{wasser_dist_cluster_2}) does allow the labelling of small clusters in $\mathcal{C}^j$, at the price of using sub-sampling to compare bigger clusters (for example more than 10000 points).
\section*{Results}
In this section we present several experiments and comparisons of our methods with other state-of-the-art procedures on a real database. 
%In this section we present several experiments and comparisons of our methods with other state of the art procedures on a real database. In Figure \ref{plot_templates} we provide visual intuition of what a cytometry looks like when understood as a mixture of normal distributions. We also provide an example of a prototype cytometry. In Figure \ref{clust_templates} we provide comparisons between the result of optimalFlowTemplates, flowMatch and what can be considered a ground truth. In Table \ref{comp_classif} we provide comparisons between state of the art unsupervised gating as given by flowMeans and our proposal for initializing \textit{tclust} with supervised information. We also provide comparisons between the state of the art supervised method deepCyTOF and our own supervised procedure optimalFlowClassification. 
\subsection*{Data}
We have a collection of 40 gated cytometries, $\mathcal{C}=\{\mathcal{C}^{1,h}, \mathcal{C}^{2,h}, \mathcal{C}^{3,h}, \mathcal{C}^{4,h}, \mathcal{C}^{5,h}, \mathcal{C}^{6,h}, \mathcal{C}^{7,h},\mathcal{C}^{8,h}, \mathcal{C}^{9,h}, \mathcal{C}^{10,h},$\\$\mathcal{C}^{11,h}, \mathcal{C}^{12,h}, \mathcal{C}^{13,h}, \mathcal{C}^{14,h}, \mathcal{C}^{15,h}, \mathcal{C}^{16,h},\mathcal{C}^{17,h}, \mathcal{C}^{18,h}, \mathcal{C}^{19,h},\mathcal{C}^{20,h}, \mathcal{C}^{21,h},\mathcal{C}^{22,s}, \mathcal{C}^{23,s}, \mathcal{C}^{24,s}, \mathcal{C}^{25,s}, \mathcal{C}^{26,s}, \mathcal{C}^{27,s}, \mathcal{C}^{28,s},$\\$\mathcal{C}^{29,h}, \mathcal{C}^{30,h}, \mathcal{C}^{31,h}, \mathcal{C}^{32,h}, \mathcal{C}^{33,h}, \mathcal{C}^{34,s},\mathcal{C}^{35,s}, \mathcal{C}^{36,h},\allowbreak \mathcal{C}^{37,h},\mathcal{C}^{38,h}, \mathcal{C}^{39,h}, \mathcal{C}^{40,h}\}$, where the super index $s$ means sick and super index $h$ means healthy. Complementary information about the cytometries can be found in Table \ref{data_info}. We are going to split them in a learning set, $\mathcal{DB}=\{\mathcal{C}^{1}, \mathcal{C}^{3},\mathcal{C}^{4},\mathcal{C}^{6},\mathcal{C}^{8},\mathcal{C}^{10},\mathcal{C}^{11},\mathcal{C}^{12},$\\$\mathcal{C}^{13},\mathcal{C}^{16},\allowbreak\mathcal{C}^{19},\mathcal{C}^{20},\mathcal{C}^{21},$ $\mathcal{C}^{22}, \allowbreak \mathcal{C}^{23},\mathcal{C}^{24},\mathcal{C}^{25},\mathcal{C}^{28},\mathcal{C}^{30},\mathcal{C}^{32},\mathcal{C}^{33},\mathcal{C}^{34},\mathcal{C}^{35},$ $\allowbreak\mathcal{C}^{36},\mathcal{C}^{37},\mathcal{C}^{38},\mathcal{C}^{39}\}$, and a test set $\mathcal{TS}=\{\mathcal{C}^{2},\mathcal{C}^{5},\mathcal{C}^{7},\mathcal{C}^{9},\allowbreak \mathcal{C}^{14},\mathcal{C}^{15},\mathcal{C}^{17},\mathcal{C}^{18},\mathcal{C}^{26},\mathcal{C}^{27},\mathcal{C}^{29},\mathcal{C}^{31},\allowbreak \mathcal{C}^{40}\}$.

Our database is formed by cytometries obtained following the Euroflow protocols using a BD FACSCanto flow cytometer in four different international centres. The size of the cytometry datasets varies from 50,000 cells to 300,000 cells. The samples are from adult male and female individuals, from a variety of age groups and with different conditions (see Table \ref{data_info}).

Clearly, there is biological variability, since there are different individuals with different conditions, ages and other different characteristics. Moreover, we have technical variability since we have different centres, different dates of measurement and different incubation times.
\begin{table*}[h!]
\begin{small}
\caption{Detailed information about the participants and the measurements for the cytometries used in the experiments. Samples come from four international centers labeled as 1 to 4. Diagnosis abbreviations correspond to: Healthy Diagnosis, Mantle Cell Lymphoma, Follicular Lymphoma, LymPhoplasmacytic Lymphoma, Chronic Lymphocytic Leukemia, Diffuse Large B-Cell Lymphoma and Hairy Cell Leukemia. The type of tested samples abbreviations correspond to: Peripherial Blood, Bone Marrow, Lymph Node. Coagulant refers to the type of coagulant used for preservation of the sample. The incubation period is measured in minutes.}\label{data_info}
\end{small}
\begin{center}
\begin{tiny}
\begin{tabular}{l|l|l|l|l|l|l|l|l|}
\cline{2-9}
                                         & \multicolumn{1}{c|}{Center} & \multicolumn{1}{c|}{Final diagnosis} & \multicolumn{1}{c|}{Tested sample} & \multicolumn{1}{c|}{\begin{tabular}[c]{@{}c@{}}Coagulant\end{tabular}} & \multicolumn{1}{c|}{Sex} & \multicolumn{1}{c|}{Age} & \multicolumn{1}{c|}{Incubation period} & \multicolumn{1}{c|}{Flow Cytometer} \\ \hline
\multicolumn{1}{|l|}{$\mathcal{C}^{1}$}  & 1                           & HD                                   & PB                                         & EDTA                                                                                                                     & M                        & 53                       & 30                                     & BD FACSCanto                                \\ \hline
\multicolumn{1}{|l|}{$\mathcal{C}^{2}$}  & 1                           & HD                                   & PB                                         & EDTA                                                                                                                     & M                        & 50                       & 30                                     & BD FACSCanto                                \\ \hline
\multicolumn{1}{|l|}{$\mathcal{C}^{3}$}  & 1                           & HD                                   & PB                                         & EDTA                                                                                                                     & M                        & 61                       & 30                                     & BD FACSCanto                                \\ \hline
\multicolumn{1}{|l|}{$\mathcal{C}^{4}$}  & 2                           & HD                                   & PB                                         & Heparin                                                                                                                  & M                        & 29                       & 30                                     & BD FACSCanto                                \\ \hline
\multicolumn{1}{|l|}{$\mathcal{C}^{5}$}  & 2                           & HD                                   & PB                                         & Heparin                                                                                                                  & M                        & 38                       & 30                                     & BD FACSCanto                                \\ \hline
\multicolumn{1}{|l|}{$\mathcal{C}^{6}$}  & 2                           & HD                                   & PB                                         & Heparin                                                                                                                  & F                        & 27                       & 30                                     & BD FACSCanto                                \\ \hline
\multicolumn{1}{|l|}{$\mathcal{C}^{7}$}  & 2                           & HD                                   & PB                                         & Heparin                                                                                                                  & F                        & NA                       & 30                                     & BD FACSCanto                                \\ \hline
\multicolumn{1}{|l|}{$\mathcal{C}^{8}$}  & 2                           & HD                                   & PB                                         & Heparin                                                                                                                  & M                        & NA                       & 30                                     & BD FACSCanto                                \\ \hline
\multicolumn{1}{|l|}{$\mathcal{C}^{9}$}  & 2                           & HD                                   & PB                                         & Heparin                                                                                                                  & F                        & NA                       & 30                                     & BD FACSCanto                                \\ \hline
\multicolumn{1}{|l|}{$\mathcal{C}^{10}$} & 2                           & HD                                   & PB                                         & Heparin                                                                                                                  & F                        & NA                       & 30                                     & BD FACSCanto                                \\ \hline
\multicolumn{1}{|l|}{$\mathcal{C}^{11}$} & 3                           & HD                                   & PB                                         & NA                                                                                                                       & M                        & 34                       & 15                                     & BD FACSCanto                                \\ \hline
\multicolumn{1}{|l|}{$\mathcal{C}^{12}$} & 3                           & HD                                   & PB                                         & NA                                                                                                                       & F                        & 33                       & 15                                     & BD FACSCanto                                \\ \hline
\multicolumn{1}{|l|}{$\mathcal{C}^{13}$} & 3                           & HD                                   & PB                                         & NA                                                                                                                       & M                        & 32                       & 15                                     & BD FACSCanto                                \\ \hline
\multicolumn{1}{|l|}{$\mathcal{C}^{14}$} & 3                           & HD                                   & PB                                         & NA                                                                                                                       & M                        & 33                       & 15                                     & BD FACSCanto                                \\ \hline
\multicolumn{1}{|l|}{$\mathcal{C}^{15}$} & 3                           & HD                                   & PB                                         & NA                                                                                                                       & F                        & 35                       & 15                                     & BD FACSCanto                                \\ \hline
\multicolumn{1}{|l|}{$\mathcal{C}^{16}$} & 3                           & HD                                   & PB                                         & EDTA                                                                                                                     & NA                       & Adult                    & 15                                     & BD FACSCanto                                \\ \hline
\multicolumn{1}{|l|}{$\mathcal{C}^{17}$} & 3                           & HD                                   & PB                                         & EDTA                                                                                                                     & NA                       & Adult                    & 15                                     & BD FACSCanto                                \\ \hline
\multicolumn{1}{|l|}{$\mathcal{C}^{18}$} & 3                           & HD                                   & PB                                         & EDTA                                                                                                                     & NA                       & Adult                    & 15                                     & BD FACSCanto                                \\ \hline
\multicolumn{1}{|l|}{$\mathcal{C}^{19}$} & 3                           & HD                                   & PB                                         & EDTA                                                                                                                     & NA                       & Adult                    & 15                                     & BD FACSCanto                                \\ \hline
\multicolumn{1}{|l|}{$\mathcal{C}^{20}$} & 3                           & HD                                   & PB                                         & EDTA                                                                                                                     & NA                       & Adult                    & 15                                     & BD FACSCanto                                \\ \hline
\multicolumn{1}{|l|}{$\mathcal{C}^{21}$} & NA                          & HD                                   & NA                                         & NA                                                                                                                       & NA                       & NA                       & NA                                     & BD FACSCanto                                \\ \hline
\multicolumn{1}{|l|}{$\mathcal{C}^{22}$} & 4                           & MCL                                  & PB                                         & NA                                                                                                                       & F                        & 82                       & 15                                     & BD FACSCanto                                \\ \hline
\multicolumn{1}{|l|}{$\mathcal{C}^{23}$} & 4                           & MCL                                  & PB                                         & NA                                                                                                                       & M                        & 70                       & 15                                     & BD FACSCanto                                \\ \hline
\multicolumn{1}{|l|}{$\mathcal{C}^{24}$} & 4                           & FL                                   & BM                                         & NA                                                                                                                       & M                        & 52                       & 15                                     & BD FACSCanto                                \\ \hline
\multicolumn{1}{|l|}{$\mathcal{C}^{25}$} & 4                           & MCL                                  & BM                                         & NA                                                                                                                       & M                        & 81                       & 15                                     & BD FACSCanto                                \\ \hline
\multicolumn{1}{|l|}{$\mathcal{C}^{26}$} & 4                           & LPL                                  & PB                                         & NA                                                                                                                       & M                        & 67                       & 15                                     & BD FACSCanto                                \\ \hline
\multicolumn{1}{|l|}{$\mathcal{C}^{27}$} & 1                           & CLL                                  & LN                                         & Other                                                                                                                    & F                        & 61                       & 30                                     & BD FACSCanto                                \\ \hline
\multicolumn{1}{|l|}{$\mathcal{C}^{28}$} & 1                           & CLL                                  & LN                                         & Other                                                                                                                    & F                        & 61                       & 30                                     & BD FACSCanto                                \\ \hline
\multicolumn{1}{|l|}{$\mathcal{C}^{29}$} & 1                           & HD                                   & PB                                         & EDTA                                                                                                                     & F                        & 27                       & 30                                     & BD FACSCanto                                \\ \hline
\multicolumn{1}{|l|}{$\mathcal{C}^{30}$} & 1                           & HD                                   & PB                                         & EDTA                                                                                                                     & M                        & 54                       & 30                                     & BD FACSCanto                                \\ \hline
\multicolumn{1}{|l|}{$\mathcal{C}^{31}$} & 1                           & HD                                   & PB                                         & EDTA                                                                                                                     & M                        & 50                       & 30                                     & BD FACSCanto                                \\ \hline
\multicolumn{1}{|l|}{$\mathcal{C}^{32}$} & 1                           & HD                                   & PB                                         & EDTA                                                                                                                     & F                        & 36                       & 30                                     & BD FACSCanto                                \\ \hline
\multicolumn{1}{|l|}{$\mathcal{C}^{33}$} & 1                           & HD                                   & PB                                         & EDTA                                                                                                                     & M                        & 74                       & 30                                     & BD FACSCanto                                \\ \hline
\multicolumn{1}{|l|}{$\mathcal{C}^{34}$} & 1                           & DLBCL                                & Other                                      & Other                                                                                                                    & M                        & 65                       & 30                                     & BD FACSCanto                                \\ \hline
\multicolumn{1}{|l|}{$\mathcal{C}^{35}$} & 1                           & HCL                                  & BM                                         & EDTA                                                                                                                     & M                        & 40                       & 30                                     & BD FACSCanto                                \\ \hline
\multicolumn{1}{|l|}{$\mathcal{C}^{36}$} & 2                           & HD                                   & PB                                         & Heparin                                                                                                                  & M                        & 38                       & 30                                     & BD FACSCanto                                \\ \hline
\multicolumn{1}{|l|}{$\mathcal{C}^{37}$} & 2                           & HD                                   & PB                                         & Heparin                                                                                                                  & F                        & 27                       & 30                                     & BD FACSCanto                                \\ \hline
\multicolumn{1}{|l|}{$\mathcal{C}^{38}$} & 2                           & HD                                   & PB                                         & Heparin                                                                                                                  & M                        & NA                       & 30                                     & BD FACSCanto                                \\ \hline
\multicolumn{1}{|l|}{$\mathcal{C}^{39}$} & 2                           & HD                                   & PB                                         & Heparin                                                                                                                  & F                        & NA                       & 30                                     & BD FACSCanto                                \\ \hline
\multicolumn{1}{|l|}{$\mathcal{C}^{40}$} & 3                           & HD                                   & PB                                         & EDTA                                                                                                                     & NA                       & NA                       & 15                                     & BD FACSCanto                                \\ \hline
\end{tabular}
\end{tiny}
\end{center}
\end{table*}
\subsection*{Measures of performance}
We need appropriate measures of the performance of the different automated gating procedures that appear in this work. We recall that we use both unsupervised and supervised methods. In this set-up an appropriate tool is the \textit{F-measure} statistic which has been used in \cite{rev_citos_1,flowMeans, flowPeaks, Li2017}. With our notation we have

\begin{equation}\label{f_measure}
F(\mathcal{C}^i,\mathcal{C}^j)=\sum_{k=1,\dots,|\mathcal{C}^i|}\frac{|\mathcal{C}^i_k|}{M}\max_{l=1,\dots,|\mathcal{C}^j|}F(\mathcal{C}^i_k,\mathcal{C}^j_l),
\end{equation}
\begin{equation}\label{f_measure_cluster}
F(\mathcal{C}^i_k,\mathcal{C}^j_l) = 2\frac{R(\mathcal{C}^i_k,\mathcal{C}^j_l)P(\mathcal{C}^i_k,\mathcal{C}^j_l)}{R(\mathcal{C}^i_k,\mathcal{C}^j_l) + P(\mathcal{C}^i_k,\mathcal{C}^j_l)},
\end{equation}
\begin{equation}\label{precision_and_recall}
R(\mathcal{C}^i_k,\mathcal{C}^j_l) = \frac{|\mathcal{C}^i_k\cap\mathcal{C}^j_l|}{|\mathcal{C}^i_k|}\quad\text{and}\quad P(\mathcal{C}^i_k,\mathcal{C}^j_l) = \frac{|\mathcal{C}^i_k\cap\mathcal{C}^j_l|}{|\mathcal{C}^j_l|}
\end{equation}
with $M= \sum_{k=1,\dots,|\mathcal{C}^i|}|\mathcal{C}^i_k|=\sum_{l=1,\dots,|\mathcal{C}^j|}|\mathcal{C}^j_l|$. We make the convention $R(\emptyset,\mathcal{C}^j_l) = P(\mathcal{C}^i_k,\emptyset) = 1$ and $R(\mathcal{C}^i_k,\emptyset) =P(\emptyset,\mathcal{C}^j_l) = 0$. Another appealing measure is the \textit{median F-measure} used in \cite{flowLearn} specifically for supervised learning. The formal definition is
\begin{align}\label{median_f_measure}
&\tilde{F}(\mathcal{C}^i,\mathcal{C}^j) =\mathrm{median}\Big\{\big\{F(\mathcal{C}^i_k,\mathcal{C}^j_{k*}):k \text{ such that } L^i_k=L^j_{k^*}\in L^i\cap L^j\big\}, \{0\}\times|L^i\triangle L^k|\Big\}&
\end{align}
where $\mathcal{C}^i$ is the considered ground truth, in our case a manual gating, and $\mathcal{C}^j$ is another classification of the same data.

To measure how similar two cytometries are, i.e., how well we do when learning from one to classify the other and how well we do when learning with the later to classify the former we introduce the following distance.
\begin{equation*}
d_{learning}(X^i,X^j)= 1-\frac{F(\mathcal{C}^j,\tilde{\mathcal{C}}^j)+F(\mathcal{C}^i,\tilde{\mathcal{C}}^i)}{2},
\end{equation*}
where $\tilde{\mathcal{C}}^j$ is the partition resulting from the classification of the data in $X^j$ using a random forest learned in $X^i$ and $\tilde{\mathcal{C}}^i$ is the partition resulting from the classification of the data in $X^i$ using a random forest learned in $X^j$. This measure gives us a notion of how close in terms of being good predictors for one another two cytometries are. We have that $0\leq d_{learning}\leq 1$, and two cytometries are interchangeable for learning if $d_{learning}$ is close to 0. When based on (\ref{median_f_measure}), we obtain a variation of this measure:
\begin{equation*}
\tilde{d}_{learning}(X^i,X^j)= 1-\frac{\tilde{F}(\mathcal{C}^j,\tilde{\mathcal{C}}^j)+\tilde{F}(\mathcal{C}^i,\tilde{\mathcal{C}}^i)}{2}.
\end{equation*}
\subsection*{Clustering cytometries and template obtention}\label{template_obtention}
Recall that $\mathcal{DB}$ is a subset of 27 cytometries taken from $\mathcal{C}$. We want to compare different methods for 
\begin{figure*}[p]
\begin{footnotesize}
\caption{Hierarchical trees for the database $\mathcal{DB}$. First row: result of optimalFlowTemplates for single linkage (left) and complete linkage (right). Second row: result of flowMatch with Mahalanobis distance (left) and Kullback-Libler divergence (right). Third row: results for $\tilde{d}_{learning}$ with single linkage (left) and complete linkage (right). Forth row: results for $d_{learning}$  with single linkage (left) and complete linkage (right).}\label{clust_templates}
\end{footnotesize}
\begin{center}
\includegraphics[scale = 0.26]{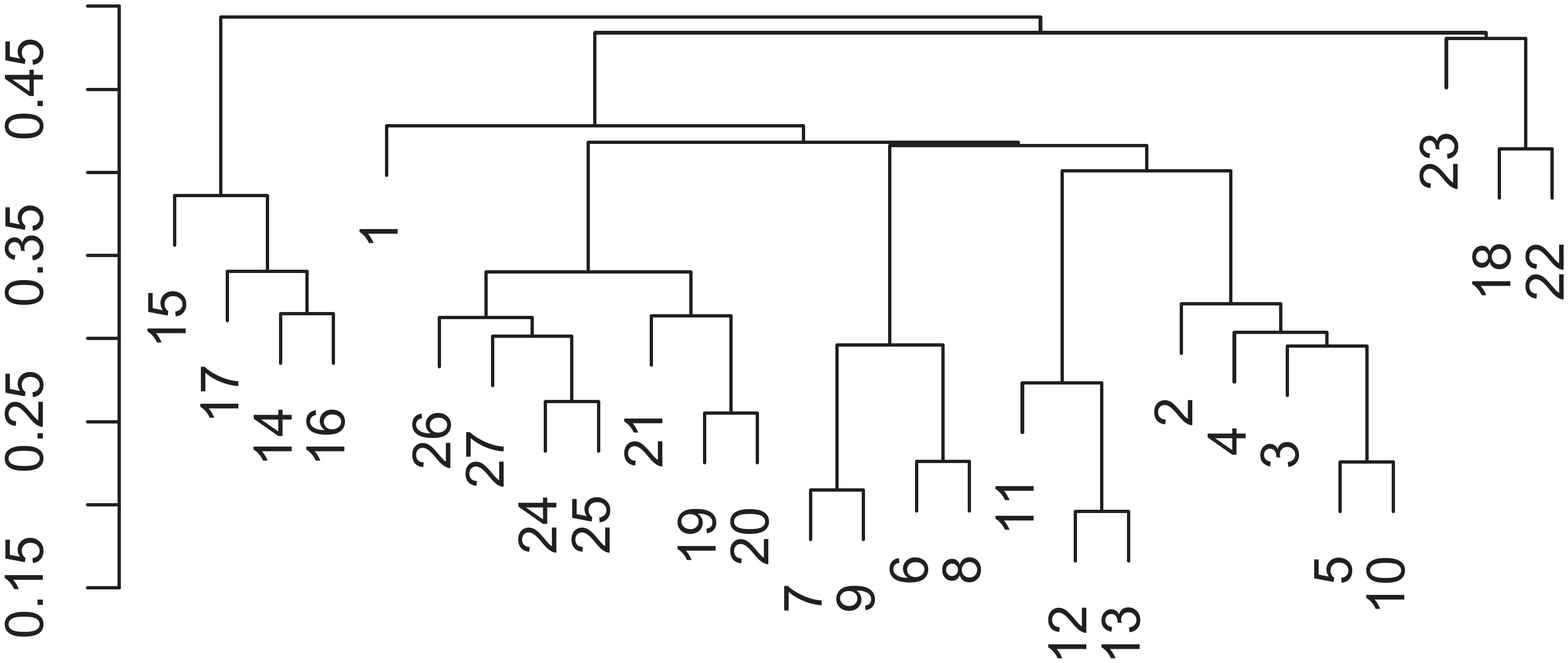}\includegraphics[scale = 0.26]{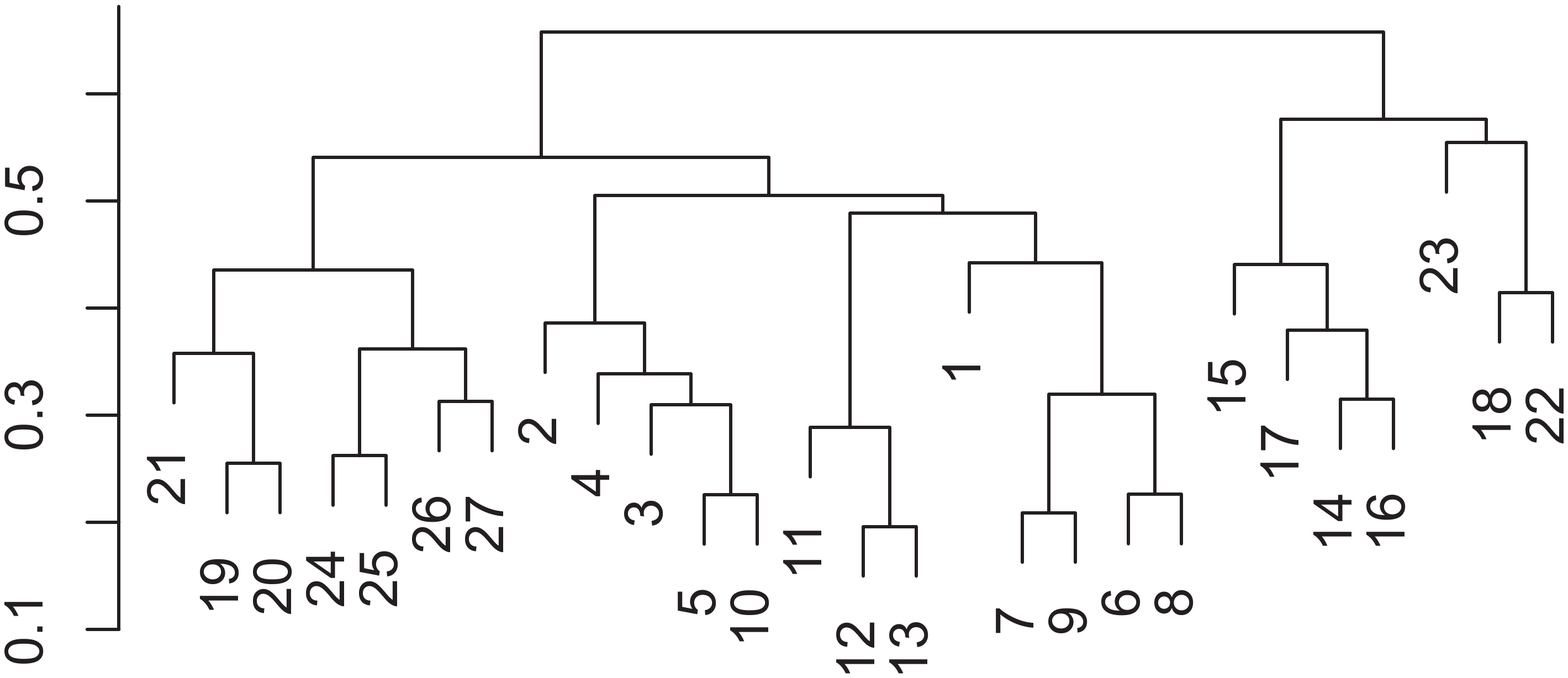}
\includegraphics[scale = 0.26]{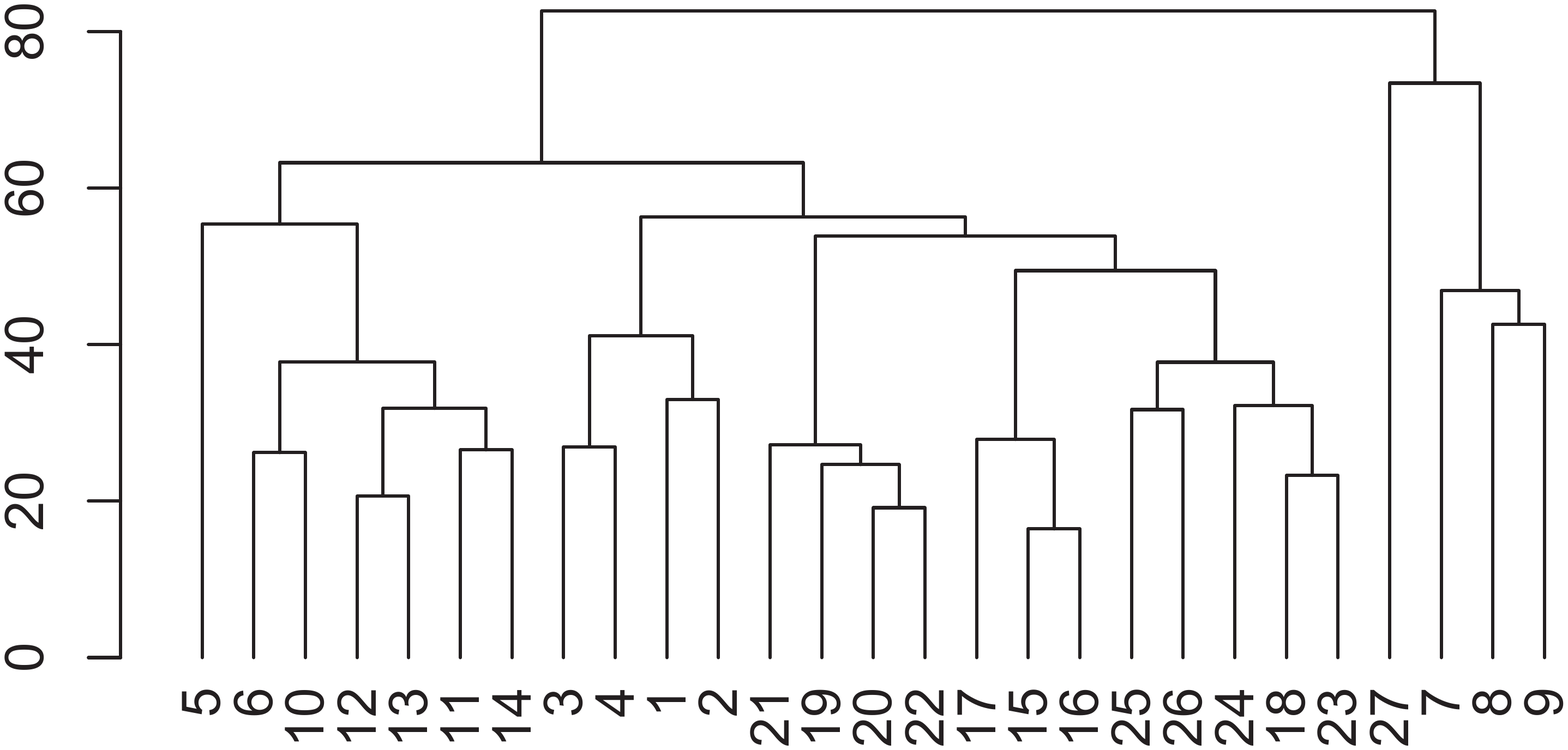}\includegraphics[scale = 0.26]{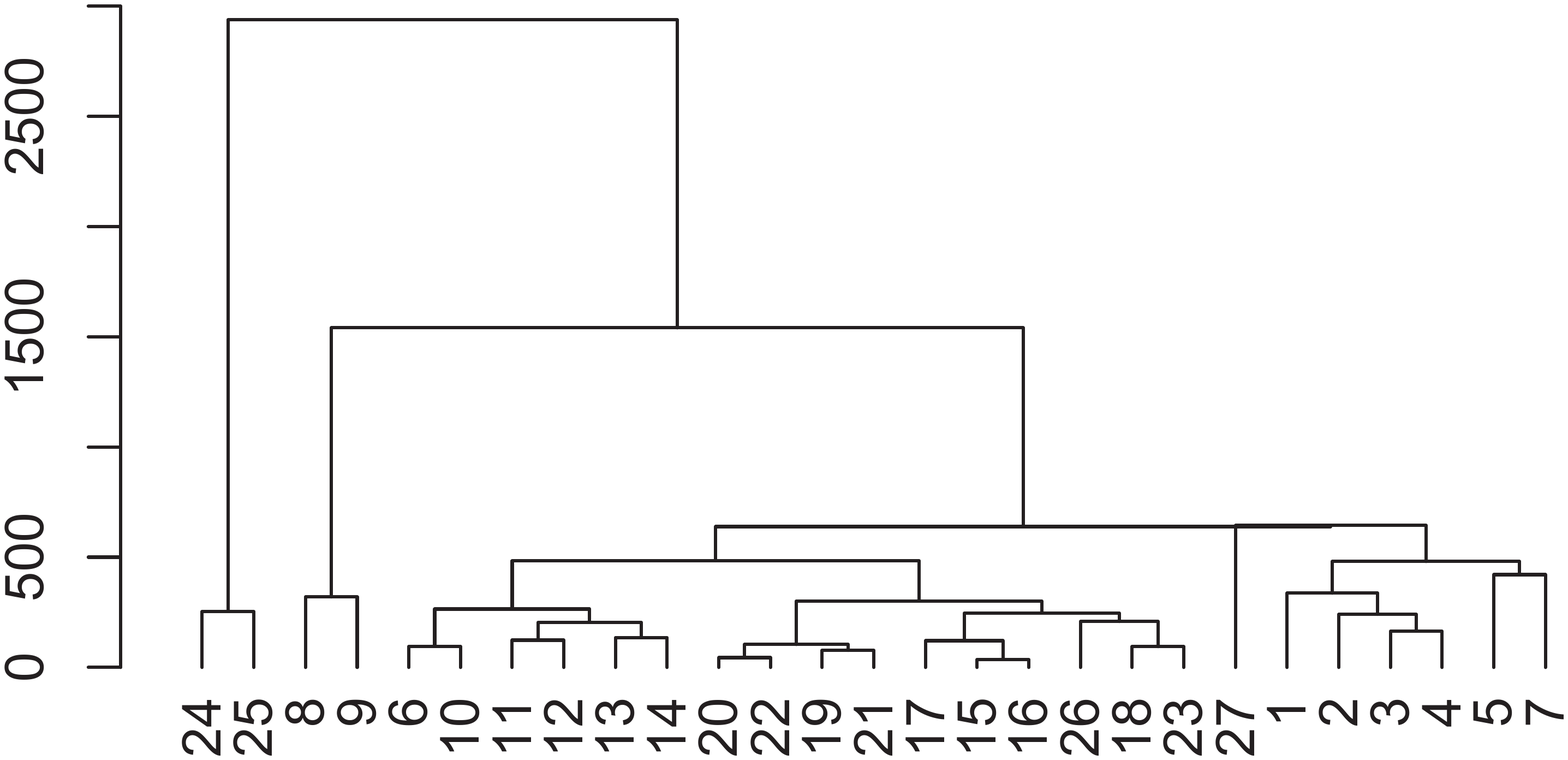}
\includegraphics[scale = 0.26]{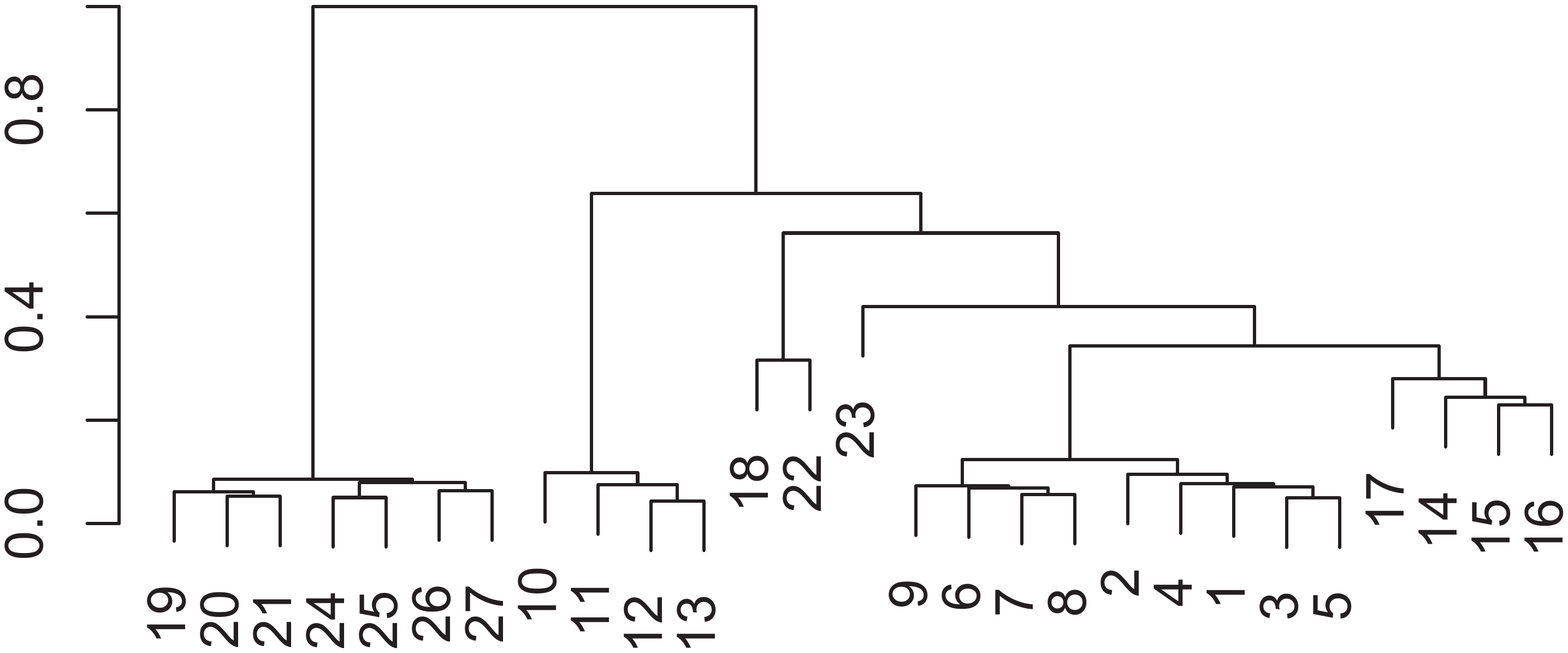}\includegraphics[scale = 0.26]{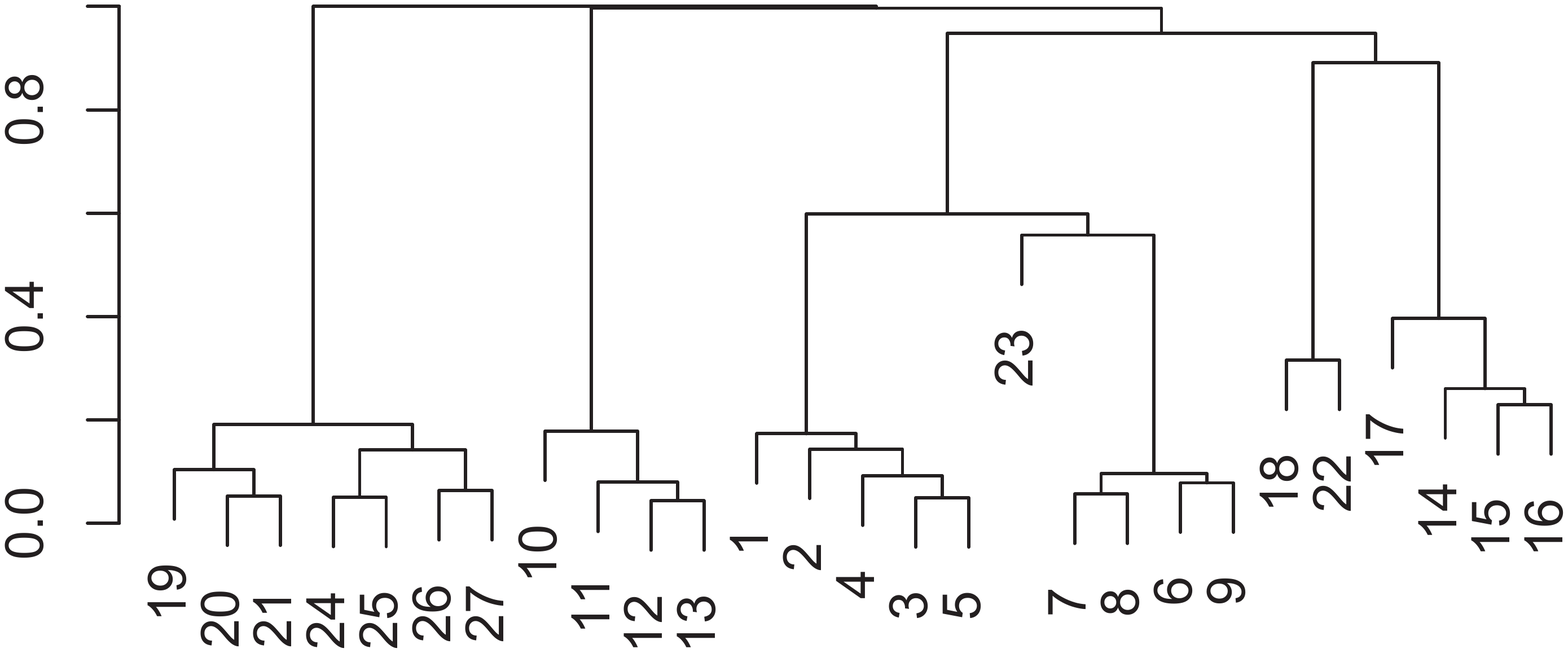}
\includegraphics[scale = 0.26]{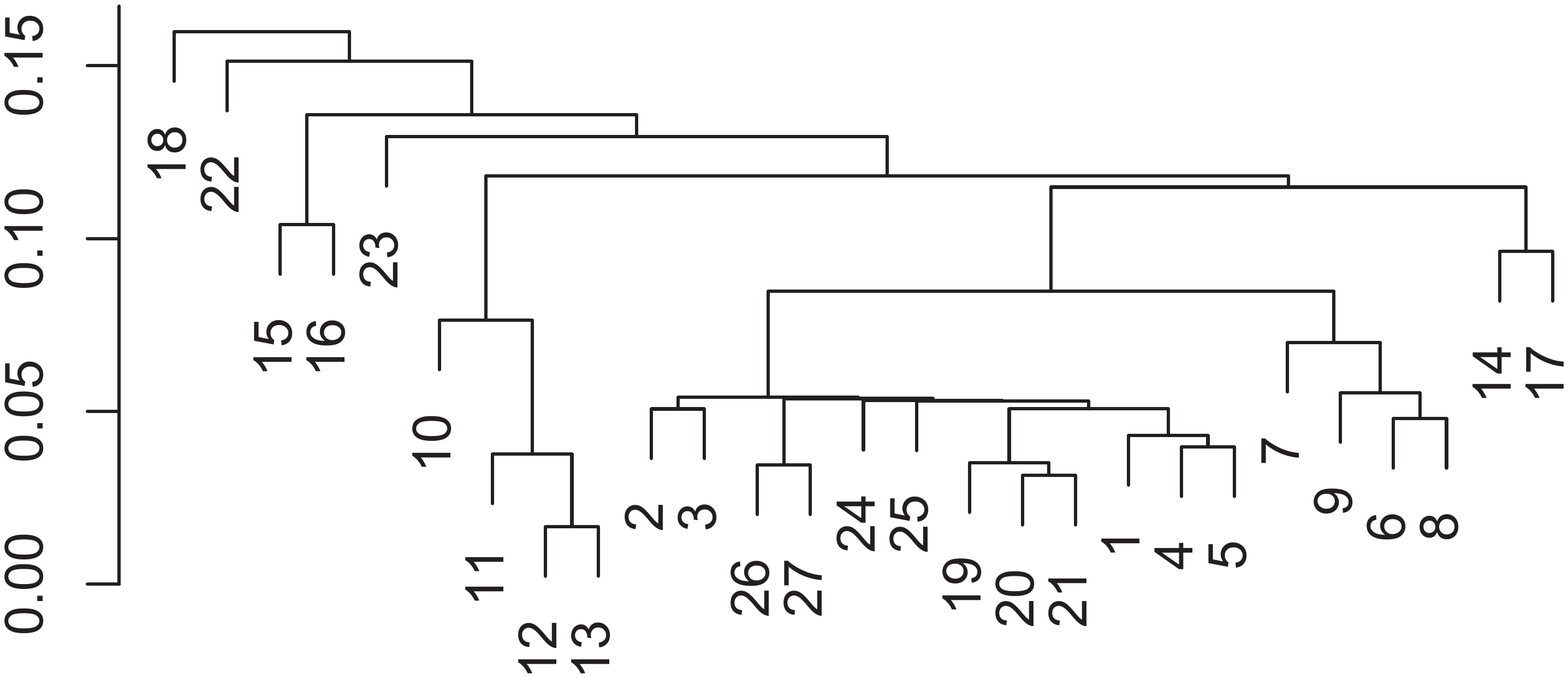}\includegraphics[scale = 0.26]{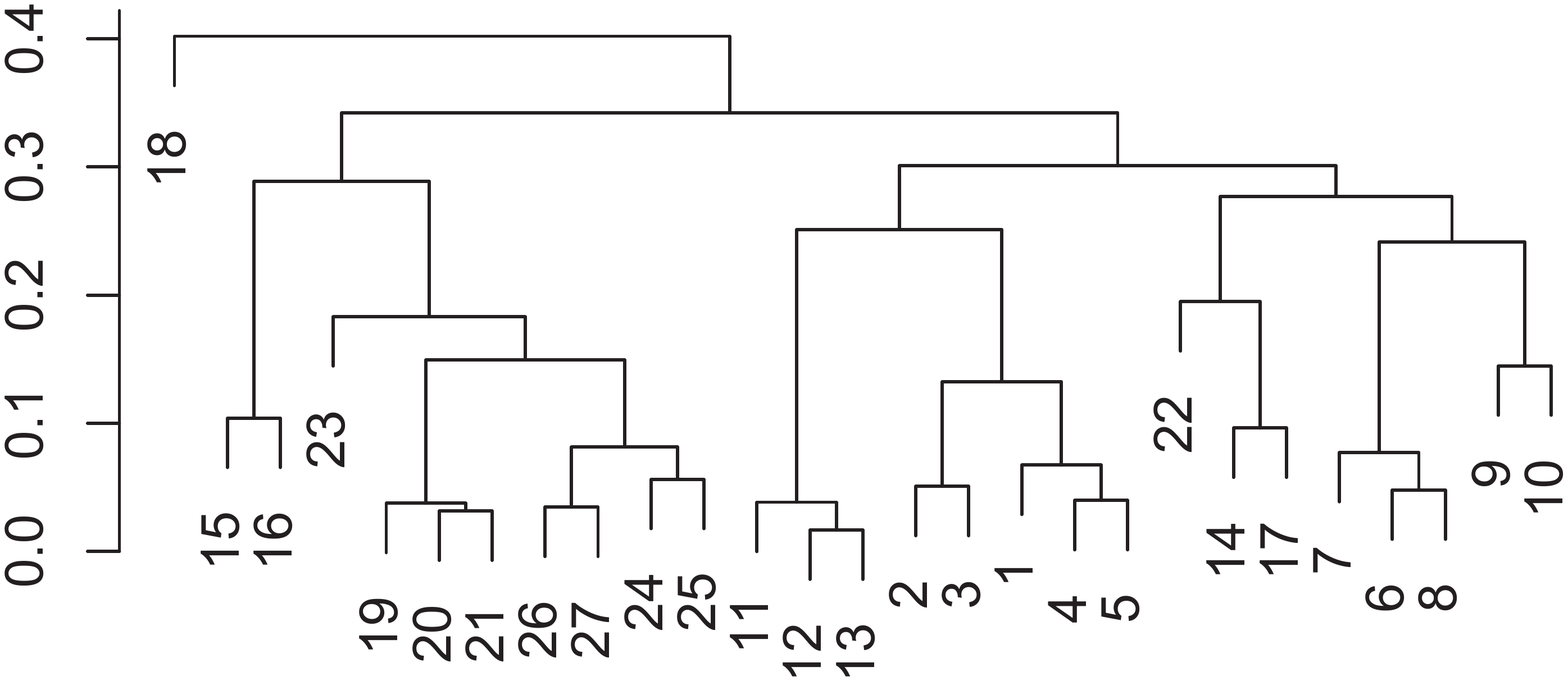}
\end{center}
\end{figure*}
clustering the database $\mathcal{DB}$. As a ground truth we use hierarchical clusterings obtained through $d_{learning}$ and $\tilde{d}_{learning}$. For a state-of-the-art comparison, we use flowMatch, described in \cite{flowMatch}. Notice that flowMatch is based in a Generalized Edge Cover procedure, a generalization of bipartite matching, where the cost between partitions is given by
\begin{equation}\label{flowMatch_KL}
d(\mathcal{C}^i,\mathcal{C}^j)=\frac{1}{k_ik_j}\sum_{k=1}^{k_i}\sum_{l=1}^{k_j}d_{KL}(\mathcal{C}^i_k,\mathcal{C}^j_l),
\end{equation}
where $d_{KL}$ is as in (\ref{dkl}), or
\begin{footnotesize}
\begin{equation}\label{flowMatch_Mahala}
d(\mathcal{C}^i,\mathcal{C}^j)=\frac{1}{k_ik_j}\sum_{k=1}^{k_i}\sum_{l=1}^{k_j}d_{Mahalanobis}(N(m^i_k, S^i_k),N(m^j_l, S^j_l)),
\end{equation}
\end{footnotesize}
where $d_{Mahalanobis}$ is the well-known Mahalanobis distance between multivariate normals.

In Figure \ref{clust_templates} we present hierarchical trees for the database $\mathcal{DB}$ that we have obtained using different clustering methods and different distances between cytometries. The top row presents the results when we use optimalFlowTemplates, or equivalently when we use similarity distance as a distance between cytometries. The second row presents the results when using flowMatch and distances between cytometries given by (\ref{flowMatch_Mahala}) and (\ref{flowMatch_KL}). The third row corresponds to using $\tilde{d}_{learning}$ as a distance between cytometries and the last row corresponds to using $d_{learning}$.

In Table \ref{database_clustering} we see two different clusterings obtained when using optimalFlowTemplates. We recall that HDBSCAN selects automatically the number of clusters. We also stress that the clustering obtained from complete linkage comes from the appropriate cut of the tree shown top right in Figure \ref{clust_templates}.

We notice that $\mathcal{DB}$ is relabelled from 1 to 27 as shown in the first row in Table  \ref{database_clustering} and these are the labels used in Figure \ref{clust_templates}. Let us stress that labels $\{14,15,16,17,18,22,23\}$ correspond to the cytometries $\{\mathcal{C}^{22}, \mathcal{C}^{23}, \mathcal{C}^{24},\mathcal{C}^{25},\mathcal{C}^{28}, \mathcal{C}^{34},\mathcal{C}^{35}\}$ that represent individuals with cancer.

\begin{table}[]
\caption{Clustering of the cytometries in $\mathcal{DB}$ obtained using optimalFlowTemplates with complete linkage hierarchical clustering looking for 7 clusters and using HDBSCAN.}\label{database_clustering}
\begin{center}
\begin{footnotesize}
\begin{tabular}{|c|c|c|c|}
\hline
 &           & \multicolumn{2}{c|}{Cluster} \\ \hline
ID & Cytometry          & Comp.-Link.   & HDBSCAN \\ \hline
1  & $\mathcal{C}^{1}$  & 1             & 1            \\ \hline
2  & $\mathcal{C}^{3}$  & 2             & 6            \\ \hline
3  & $\mathcal{C}^{4}$  & 2             & 6            \\ \hline
4  & $\mathcal{C}^{6}$  & 2             & 6            \\ \hline
5  & $\mathcal{C}^{8}$  & 2             & 6            \\ \hline
6  & $\mathcal{C}^{10}$ & 1             & 8            \\ \hline
7  & $\mathcal{C}^{11}$ & 1             & 9            \\ \hline
8  & $\mathcal{C}^{12}$ & 1             & 8            \\ \hline
9  & $\mathcal{C}^{13}$ & 1             & 9            \\ \hline
10 & $\mathcal{C}^{16}$ & 3             & 6            \\ \hline
11 & $\mathcal{C}^{19}$ & 3             & 7            \\ \hline
12 & $\mathcal{C}^{20}$ & 3             & 7            \\ \hline
13 & $\mathcal{C}^{21}$ & 4             & 7            \\ \hline
14 & $\mathcal{C}^{22}$ & 4             & 2            \\ \hline
15 & $\mathcal{C}^{23}$ & 4             & 2            \\ \hline
16 & $\mathcal{C}^{24}$ & 4             & 2            \\ \hline
17 & $\mathcal{C}^{25}$ & 4             & 2            \\ \hline
18 & $\mathcal{C}^{28}$ & 5             & 3            \\ \hline
19 & $\mathcal{C}^{30}$ & 6             & 4            \\ \hline
20 & $\mathcal{C}^{32}$ & 6             & 4            \\ \hline
21 & $\mathcal{C}^{33}$ & 6             & 4            \\ \hline
22 & $\mathcal{C}^{34}$ & 5             & 3            \\ \hline
23 & $\mathcal{C}^{35}$ & 7             & 3            \\ \hline
24 & $\mathcal{C}^{36}$ & 6             & 5            \\ \hline
25 & $\mathcal{C}^{37}$ & 6             & 5            \\ \hline
26 & $\mathcal{C}^{38}$ & 6             & 5            \\ \hline
27 & $\mathcal{C}^{39}$ & 6             & 5            \\ \hline
\end{tabular}
\end{footnotesize}
\end{center}
\end{table}

%Once we have seen the intuitive meaning of clustering cytometries and the obtention of templates, we want to compare different methods to cluster the database $\mathcal{DB}$. We use for a ground truth the simple linkage hierarchical clusterings obtained using $d_{learning}$, bottom right of Figure \ref{clust_templates}, and using $\tilde{d}_{learning}$, bottom left of Figure \ref{clust_templates}. For a state of the art comparison, we use flowMatch, described in \cite{flowMatch}, using Mahalanobis distance, depicted top right in Figure \ref{clust_templates}. The clustering obtained by optimalFlowTemplates when using single linkage hierarchical clustering is shown top left in Figure \ref{clust_templates}. More comparisons can be seen in Figure \ref{clust_templates_2} in the Supplementary material. At a first glance it is clear that the results form optimalFlowTemplates are much more similar to the ground truth than those of flowMatch.  
\subsection*{Gating and classification}\label{results_gating}
We will use the results of optimalFlowTemplates applied to the database $\mathcal{DB}$, introduced in the previous section, as entries to optimalFlowClassification in order to do gating in $\mathcal{TS}$. For the cytometries in $\mathcal{TS}$, we also perform an unsupervised gating given by flowMeans. Results are shown in the first and last columns of Table \ref{comp_classif}.

We also want to compare with a state-of-the-art supervised procedure. In this case we will use deepCyTOF, with some bug corrections and some adaptations to our setting of the github version, implemented in Python with \textit{tensorflow} 0.12 and \textit{keras} 1.2.2. In order to use deepCyTOF we need cytometries with the same number and type of cell types so we use a data set $\mathcal{TS}' = \{\mathcal{C}^{2'},\mathcal{C}^{5'},\mathcal{C}^{7'},\mathcal{C}^{14'},\mathcal{C}^{15'},\mathcal{C}^{17'},\mathcal{C}^{18'}\}$, where we have extracted the common groups from the original cytometries. Hence comparisons in Table \ref{comp_classif} are biased in favour of deepCyTOF, since for optimalFlowClassification we use the original complete cytometries. We want to emphasize that deepCyTOF only uses the supervised information of one of the cytometries in $\mathcal{TS}'$ to classify all other members. This is shown in Table \ref{comp_classif} in italic. Results of deepCyTOF are provided, with domain adaptation and without de-noising, since all entries are classified, in column 2 of Table \ref{comp_classif}.

\begin{table*}[h!]
\caption{Table of F-measure statistics as given by (\ref{f_measure}), where we use the manual gating as the ground truth. First column: results of the unsupervised gating procedure flowMeans on $\mathcal{TS}$. Second column: results of the supervised procedure deepCyTOF on $\mathcal{TS}'$. Third column, results of deepCyTOF on the clusters $\mathcal{TS}'_1$, $\mathcal{TS}'_2$ and $\mathcal{TS}'_3$ produced by optimalFlowTemplates. Forth column: results of our supervised procedure optimalFlowTemplates + optimalFlowClassification on $\mathcal{TS}$. In bold we have the best performance according to F-measure.} \label{comp_classif}
\begin{center}
\begin{small}
\begin{tabular}{l|l|ll|l|}
\cline{2-5}
                                       & flowMeans & \multicolumn{1}{l|}{deepCyTOF} & \begin{tabular}[c]{@{}l@{}}optimalFlowTemplates \\ + deepCyTOF\end{tabular} & \begin{tabular}[c]{@{}l@{}}optimalFlowTemplates \\ + optimalFlowClassification\end{tabular} \\ \hline
\multicolumn{1}{|c|}{$\mathcal{C}^2$}  & \multicolumn{1}{c|}{0.8988}   & \multicolumn{1}{c|}{0.9546}    & \multicolumn{1}{c|}{\textbf{\emph{0.9736}}}                                                                    & \multicolumn{1}{c|}{0.9610}                                                                                      \\ \hline
\multicolumn{1}{|c|}{$\mathcal{C}^5$}  & \multicolumn{1}{c|}{0.8977}    & \multicolumn{1}{c|}{0.9161}    & \multicolumn{1}{c|}{0.9196}                                                                      & \multicolumn{1}{c|}{\textbf{0.9587}}                                                                                      \\ \hline
\multicolumn{1}{|c|}{$\mathcal{C}^7$}  & \multicolumn{1}{c|}{0.9508}    & \multicolumn{1}{c|}{0.7514}    & \multicolumn{1}{c|}{\emph{\textbf{0.9769}}}                                                                      & \multicolumn{1}{c|}{0.9768}                                                                                      \\ \hline
\multicolumn{1}{|c|}{$\mathcal{C}^9$}  & \multicolumn{1}{c|}{0.8936}    &                                &                                                                             & \multicolumn{1}{c|}{\textbf{0.9172}}                                                                                      \\ \hline
\multicolumn{1}{|c|}{$\mathcal{C}^{14}$} & \multicolumn{1}{c|}{0.9004}    & \multicolumn{1}{c|}{\emph{\textbf{0.9838}}}    & \multicolumn{1}{c|}{0.9530}                                                                      & \multicolumn{1}{c|}{0.9066}                                                                                      \\ \hline
\multicolumn{1}{|c|}{$\mathcal{C}^{15}$} & \multicolumn{1}{c|}{0.8974}    & \multicolumn{1}{c|}{0.9408}    & \multicolumn{1}{c|}{0.9352}                                                                      & \multicolumn{1}{c|}{\textbf{0.9556}}                                                                                      \\ \hline
\multicolumn{1}{|c|}{$\mathcal{C}^{17}$} & \multicolumn{1}{c|}{0.9405}    & \multicolumn{1}{c|}{0.7847}    & \multicolumn{1}{c|}{\emph{0.9810}}                                                                      & \multicolumn{1}{c|}{\textbf{0.9848}}                                                                                      \\ \hline
\multicolumn{1}{|c|}{$\mathcal{C}^{18}$} & \multicolumn{1}{c|}{0.9004}    & \multicolumn{1}{c|}{0.7837}    & \multicolumn{1}{c|}{0.9796                                                       } & \multicolumn{1}{c|}{\textbf{0.9849}}                                                                                      \\ \hline
\multicolumn{1}{|c|}{$\mathcal{C}^{26}$} & \multicolumn{1}{c|}{0.9024}    &                                &                                                                             & \multicolumn{1}{c|}{\textbf{0.9313}}                                                                                      \\ \cline{1-2} \cline{5-5} 
\multicolumn{1}{|c|}{$\mathcal{C}^{27}$} & \multicolumn{1}{c|}{0.8645}    &                                &                                                                             & \multicolumn{1}{c|}{\textbf{0.9306}}                                                                                      \\ \cline{1-2} \cline{5-5} 
\multicolumn{1}{|c|}{$\mathcal{C}^{29}$} & \multicolumn{1}{c|}{0.9475}    &                                &                                                                             & \multicolumn{1}{c|}{\textbf{0.9744}}                                                                                      \\ \cline{1-2} \cline{5-5} 
\multicolumn{1}{|c|}{$\mathcal{C}^{31}$} & \multicolumn{1}{c|}{0.9290}    &                                &                                                                             & \multicolumn{1}{c|}{\textbf{0.9656}}                                                                                      \\ \cline{1-2} \cline{5-5} 
\multicolumn{1}{|c|}{$\mathcal{C}^{40}$} & \multicolumn{1}{c|}{0.9330}    &                                &                                                                             & \multicolumn{1}{c|}{\textbf{0.9538}}                                                                                      \\ \cline{1-2} \cline{5-5} 
\end{tabular}
\end{small}
\end{center}
\end{table*}
%
%However, we also want to compare with a state of the art supervised procedure. In this case we will use deepCyTOF, with some bug corrections and some adaptations to our setting of the github version, implemented in Python with \textit{tensorflow} 0.12 and \textit{keras} 1.2.2. In order to use deepCyTOF we need cytometries with the same number and type of cell types so we use a data set $\mathcal{TS}' = \{\tilde{\mathcal{C}}^2,\tilde{\mathcal{C}}^5,\tilde{\mathcal{C}}^7,\tilde{\mathcal{C}}^{14},\tilde{\mathcal{C}}^{15},\tilde{\mathcal{C}}^{17},\tilde{\mathcal{C}}^{18}\}$, where we have extracted the common groups from the original cytometries. We recall that deepCyTOF only uses the supervised information of one of the cytometries in $\mathcal{TS}'$ to classify all other members. We see the results of deepCyTOF, with domain adaptation and without de-noising, since all entries are classified, in column 1 of Table \ref{comp_classif}. DeepCyTOF's performance is rather poor, achieving worst F-measure than flowMeans in 6 of the 9 cases and also for all applicable cases (cytometries 1,18,20) than optimalFlowTemplates+optimalFlowClassification.

\begin{figure}[h]
\begin{center}
\includegraphics[scale = 0.26]{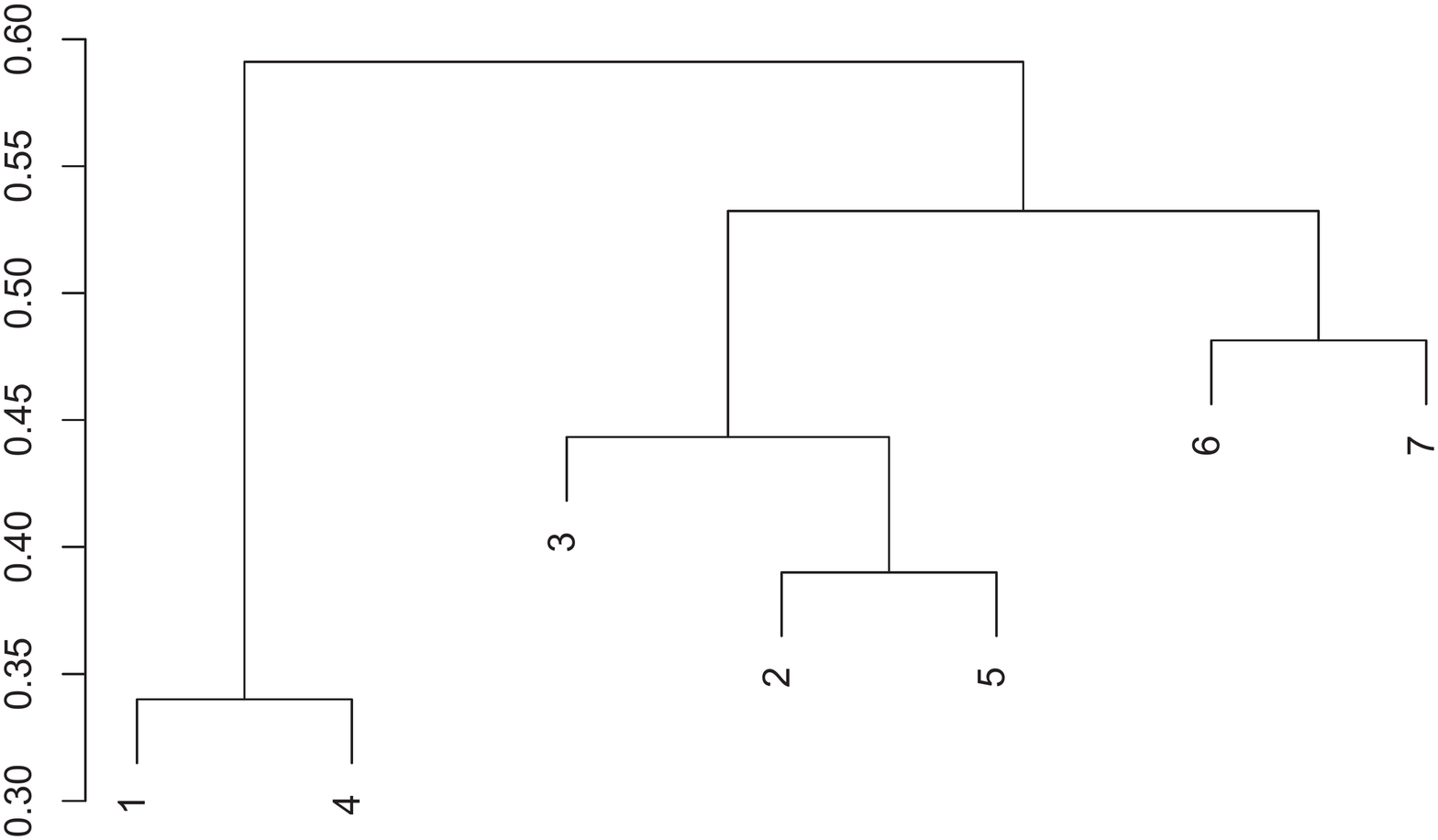}
\end{center}
\caption{Hierarchical tree obtained by using optimalFlowTemplates with complete linkage on the databse $\mathcal{TS}'$ after gating each cytometry with the unsupervised procedure flowMeans.}\label{clust_deepcytof}
\end{figure}

We see that deeepCyTOF performs rather poorly for $\{\mathcal{C}^{7'},\mathcal{C}^{17'},\mathcal{C}^{18'}\}$, this is due to the high variability of the cytometries in $\mathcal{TS}'$ that can not be accommodated by the domain adaptation procedure of deepCyTOF. Hence if we were able to reduce this variability, deepCyTOF should give better results. Indeed, if we use flowMeans to gate the cytometries in $\mathcal{TS}'$, and then we use optimalFlowTemplates, we obtain the hierarchical tree presented in Figure \ref{clust_deepcytof}. It suggests splitting $\mathcal{TS}'$ into $\mathcal{TS}'_1=\{\mathcal{C}^{2'},\mathcal{C}^{14'}\}$, $\mathcal{TS}'_2=\{\mathcal{C}^{5'},\mathcal{C}^{7'},\mathcal{C}^{15'}\}$ and $\mathcal{TS}'_3=\{\mathcal{C}^{17'},\mathcal{C}^{18'}\}$. We highlight that until now we have not used supervised information. Applying deepCyTOF to $\mathcal{TS}'_1$, $\mathcal{TS}'_2$ and $\mathcal{TS}'_3$ we obtain the results in column 3 of Table \ref{comp_classif}. Again, in italics we have the cytometry which deepCyTOF uses for learning in each group.

\begin{table*}[h!]
\begin{footnotesize}
\caption{Parameters and performance (as measured by (\ref{median_f_measure}) and manual gating as ground truth) of the best results obtained by optimalFlowTemplates + optimalFlowClassification on $\mathcal{TS}$. \emph{Database clustering} refers to the clustering method used in line 17 in Algorithm \ref{optimalFlowTemplates}. \emph{Template Formation} refers to the method used in line 19 in Algorithm \ref{optimalFlowTemplates}. \emph{Assigned Cluster} refers to the label of the cluster as given in Table \ref{database_clustering} to which the new cytometry is assigned. \emph{Sample Clustering} refers to how we obtain $\mathcal{C}^u$ in Algorithm \ref{optimalFlowClassification}. \emph{Supervised Classification} refers to the method used in line 13 in Algorithm \ref{optimalFlowClassification}. \emph{Assigned Cytometry} refers to the optimal cytometry in the respective cluster that is used for learning (when applicable).}\label{table_medain_f}
\end{footnotesize}
\begin{center}
\begin{footnotesize}
\begin{tabular}{l|l|ll}
\cline{2-4}
                                                         & \color{red}{$\mathcal{C}^2$}  & \multicolumn{1}{l|}{\color{red}{$\mathcal{C}^{5}$}}                                                                       & \multicolumn{1}{l|}{\color{red}{$\mathcal{C}^{7}$}}                          \\ \hline
\multicolumn{1}{|l|}{\textit{Median F- measure}}         & 0.9441931                   & \multicolumn{1}{l|}{0.8530806}                                                                                        & \multicolumn{1}{l|}{0.957045}                                            \\ \hline
\multicolumn{1}{|l|}{\textit{Database Clustering}}       & Complete Linkage            & \multicolumn{1}{l|}{HDBSCAN}                                                                                          & \multicolumn{1}{l|}{Complete Linkage}                                    \\ \hline
\multicolumn{1}{|l|}{\textit{Template Formation}}        & Pooling                     & \multicolumn{1}{l|}{Pooling}                                                                                          & \multicolumn{1}{l|}{HDBSCAN}                                             \\ \hline
\multicolumn{1}{|l|}{\textit{Assigned Cluster}}          & 1                           & \multicolumn{1}{l|}{6}                                                                                                & \multicolumn{1}{l|}{2}                                                   \\ \hline
\multicolumn{1}{|l|}{\textit{Sample Clustering}}         & tclust                      & \multicolumn{1}{l|}{tclust}                                                                                           & \multicolumn{1}{l|}{tclust}                                              \\ \hline
\multicolumn{1}{|l|}{\textit{Supervised Classification}} & QDA               & \multicolumn{1}{l|}{\begin{tabular}[c]{@{}l@{}}QDA from Template\end{tabular}} & \multicolumn{1}{l|}{Random Forest}                                       \\ \hline
\multicolumn{1}{|l|}{\textit{Assigned Cytometry}}        & $\mathcal{C}^{1}$           & \multicolumn{1}{l|}{}                                                                                                 & \multicolumn{1}{l|}{$\mathcal{C}^{8}$}                                   \\ \hline
                                                         & \color{red}{$\mathcal{C}^{9}$}  & \multicolumn{1}{l|}{\color{red}{$\mathcal{C}^{14}$}}                                                                      & \multicolumn{1}{l|}{\color{red}{$\mathcal{C}^{15}$}}                         \\ \hline
\multicolumn{1}{|l|}{\textit{Median F- measure}}         & 0.9458429                   & \multicolumn{1}{l|}{0.9254252}                                                                                        & \multicolumn{1}{l|}{0.8807339}                                           \\ \hline
\multicolumn{1}{|l|}{\textit{Database Clustering}}       & HDBSCAN                     & \multicolumn{1}{l|}{HDBSCAN}                                                                                          & \multicolumn{1}{l|}{HDBSCAN}                                             \\ \hline
\multicolumn{1}{|l|}{\textit{Template Formation}}        & Pooling                     & \multicolumn{1}{l|}{k-barycenter}                                                                                     & \multicolumn{1}{l|}{k-barycenter}                                        \\ \hline
\multicolumn{1}{|l|}{\textit{Assigned Cluster}}          & 9                           & \multicolumn{1}{l|}{1}                                                                                                & \multicolumn{1}{l|}{1}                                                   \\ \hline
\multicolumn{1}{|l|}{\textit{Sample Clustering}}         & tclust                      & \multicolumn{1}{l|}{tclust}                                                                                           & \multicolumn{1}{l|}{tclust}                                           \\ \hline
\multicolumn{1}{|l|}{\textit{Supervised Classification}} & QDA                         & \multicolumn{1}{l|}{Label transfer with (\ref{wasser_dist_cluster_2})}                                              & \multicolumn{1}{l|}{Random Forest}                                       \\ \hline
\multicolumn{1}{|l|}{\textit{Assigned Cytometry}}        & $\mathcal{C}^{13}$          & \multicolumn{1}{l|}{}                                                                                                 & \multicolumn{1}{l|}{$\mathcal{C}^{1}$}                                  \\ \hline
                                                         & \color{red}{$\mathcal{C}^{17}$} & \multicolumn{1}{l|}{\color{red}{$\mathcal{C}^{18}$}}                                                                      & \multicolumn{1}{l|}{\color{red}{$\mathcal{C}^{26}$}}                         \\ \hline
\multicolumn{1}{|l|}{\textit{Median F- measure}}         & 0.9679446                   & \multicolumn{1}{l|}{0.9575489}                                                                                        & \multicolumn{1}{l|}{0.8316279}                                           \\ \hline
\multicolumn{1}{|l|}{\textit{Database Clustering}}       & HDBSCAN                     & \multicolumn{1}{l|}{HDBSCAN}                                                                                          & \multicolumn{1}{l|}{Complete Linkage}                                    \\ \hline
\multicolumn{1}{|l|}{\textit{Template Formation}}        & HDBSCAN                     & \multicolumn{1}{l|}{HDBSCAN}                                                                                          & \multicolumn{1}{l|}{HDBSCAN}                                             \\ \hline
\multicolumn{1}{|l|}{\textit{Assigned Cluster}}          & 7                           & \multicolumn{1}{l|}{7}                                                                                                & \multicolumn{1}{l|}{4}                                                   \\ \hline
\multicolumn{1}{|l|}{\textit{Sample Clustering}}         & tclust                      & \multicolumn{1}{l|}{flowMeans}                                                                                        & \multicolumn{1}{l|}{tclust}                                              \\ \hline
\multicolumn{1}{|l|}{\textit{Supervised Classification}} & Random Forest               & \multicolumn{1}{l|}{Random Forest}                                                                                    & \multicolumn{1}{l|}{Random Forest} \\ \hline
\multicolumn{1}{|l|}{\textit{Assigned Cytometry}}        & $\mathcal{C}^{20}$          & \multicolumn{1}{l|}{$\mathcal{C}^{20}$}                                                                               & \multicolumn{1}{l|}{$\mathcal{C}^{24}$}                                                    \\ \hline
                                                         & \color{red}{$\mathcal{C}^{27}$} & \multicolumn{1}{l|}{\color{red}{$\mathcal{C}^{29}$}}                                                                      & \multicolumn{1}{l|}{\color{red}{$\mathcal{C}^{31}$}}                         \\ \hline
\multicolumn{1}{|l|}{\textit{Median F- measure}}         & 0.9312977                   & \multicolumn{1}{l|}{0.9259644}                                                                                        & \multicolumn{1}{l|}{0.931515}                                            \\ \hline
\multicolumn{1}{|l|}{\textit{Database Clustering}}       & Complete Linkage            & \multicolumn{1}{l|}{Complete Linkage}                                                                                 & \multicolumn{1}{l|}{HDBSCAN}                                             \\ \hline
\multicolumn{1}{|l|}{\textit{Template Formation}}        & Pooling                     & \multicolumn{1}{l|}{k-barycenter}                                                                                     & \multicolumn{1}{l|}{Pooling}                                             \\ \hline
\multicolumn{1}{|l|}{\textit{Assigned Cluster}}          & 5                           & \multicolumn{1}{l|}{6}                                                                                                & \multicolumn{1}{l|}{4}                                                   \\ \hline
\multicolumn{1}{|l|}{\textit{Sample Clustering}}         & tclust                      & \multicolumn{1}{l|}{flowMeans}                                                                                        & \multicolumn{1}{l|}{tclust}                                              \\ \hline
\multicolumn{1}{|l|}{\textit{Supervised Classification}} & Random Forest               & \multicolumn{1}{l|}{Random Forest}                                                                                    & \multicolumn{1}{l|}{QDA from Template}                                   \\ \hline
\multicolumn{1}{|l|}{\textit{Assigned Cytometry}}        & $\mathcal{C}^{28}$          & \multicolumn{1}{l|}{$\mathcal{C}^{33}$}                                                                               & \multicolumn{1}{l|}{}                                                    \\ \hline
                                                         & \color{red}{$\mathcal{C}^{40}$} &                                                                                                                       &                                                                          \\ \cline{1-2}
\multicolumn{1}{|l|}{\textit{Median F- measure}}         & 0.8240522                   &                                                                                                                       &                                                                          \\ \cline{1-2}
\multicolumn{1}{|l|}{\textit{Database Clustering}}       & Complete Linkage            &                                                                                                                       &                                                                          \\ \cline{1-2}
\multicolumn{1}{|l|}{\textit{Template Formation}}        & Pooling                     &                                                                                                                       &                                                                          \\ \cline{1-2}
\multicolumn{1}{|l|}{\textit{Assigned Cluster}}          & 6                           &                                                                                                                       &                                                                          \\ \cline{1-2}
\multicolumn{1}{|l|}{\textit{Sample Clustering}}         & tclust                      &                                                                                                                       &                                                                          \\ \cline{1-2}
\multicolumn{1}{|l|}{\textit{Supervised Classification}} & Random Forest               &                                                                                                                       &                                                                          \\ \cline{1-2}
\multicolumn{1}{|l|}{\textit{Assigned Cytometry}}        & $\mathcal{C}^{30}$          &                                                                                                                       &                                                                          \\ \cline{1-2}
\end{tabular}
\end{footnotesize}
\end{center}
\end{table*}

We present the best results, as measured by median F-measure (\ref{median_f_measure}), of applying optimalFlowTemplates + optimalFlowClassification to $\mathcal{TS}$ in Table \ref{table_medain_f}. For instance, for $\mathcal{C}^{27}$, a 61 year old female with Chronic Lymphocytic Leukemia (see Table \ref{data_info}), we have obtained a very satisfactory median F-measure of 0.9313. This value has been obtained through the use of optimalFlowTemplate and then through optimalFlowClassification. What follows is an extended explanation of how to understand the entries of Table \ref{table_medain_f}.
For optimalFlowTemplates we have used complete linkage to cluster the cytometries in $\mathcal{DB}$ (third column in Table \ref{database_clustering}) and pooling to obtain the templates. This information is provided in the entries \emph{Database Clustering} and \emph{Template Formation} in Table \ref{table_medain_f} corresponding to $\mathcal{C}^{27}$ (in red). For optimalFlowClassification, we have clustered $\mathcal{C}^{27}$ (without using the manual gating information) with tclust as indicated in lines 1-5 in Algorithm \ref{optimalFlowClassification} obtaining $\mathcal{C}^u$. This is reflected in the entry \emph{Sample Clustering}. Then, we have assigned $\mathcal{C}^{u}$ to the cluster, $\{\mathcal{C}^{28},\mathcal{C}^{34}\}$, labelled as 5, in column three, in Table \ref{database_clustering}. This is shown in the entry \emph{Assigned Cluster}. Since we want to use Random Forest as a classification method, as reflected in the entry \emph{Supervised Classification}, we have assigned $\mathcal{C}^u$ to the closest cytometry in similarity distance in the cluster, i.e, the assigned cytometry for learning is $\mathcal{C}^{28}$ as reflected in the entry \emph{Assigned Cytometry}.

\begin{table*}[h!]
\caption{Table giving the F-measure values (as in (\ref{f_measure_cluster})) for each cell type for $\mathcal{C}^{17}$ and $\mathcal{C}^{27}$ which are used to obtain the respective median F-measure value in Table \ref{table_medain_f}. Recall and precision are defined in (\ref{precision_and_recall}).}
\label{f_measure_cell_types}
\begin{small}
\begin{tabular}{c|ccc|ccc}
\cline{2-7}
                                         & \multicolumn{3}{c|}{$\mathcal{C}^{17}$}                                 & \multicolumn{3}{c|}{$\mathcal{C}^{27}$}                                                      \\ \cline{2-7} 
                                         & \multicolumn{1}{c|}{F-measure} & \multicolumn{1}{c|}{Precision} & Recall & \multicolumn{1}{c|}{F-measure} & \multicolumn{1}{c|}{Precision} & \multicolumn{1}{c|}{Recall} \\ \hline
\multicolumn{1}{|c|}{Abnormal Sig Kappa} &                               &                                &        & \multicolumn{1}{c|}{0.9697}   & \multicolumn{1}{c|}{0.9478}    & \multicolumn{1}{c|}{0.9925} \\ \hline
\multicolumn{1}{|c|}{CD4+CD8-}           & \multicolumn{1}{c|}{0.9973}   & \multicolumn{1}{c|}{0.9982}    & 0.9965 & \multicolumn{1}{c|}{0.9828}   & \multicolumn{1}{c|}{0.9796}    & \multicolumn{1}{c|}{0.9859} \\ \hline
\multicolumn{1}{|c|}{CD8+CD4-}           & \multicolumn{1}{c|}{0.9960}   & \multicolumn{1}{c|}{0.9960}    & 0.9960 & \multicolumn{1}{c|}{0.9769}   & \multicolumn{1}{c|}{0.9835}    & \multicolumn{1}{c|}{0.9704} \\ \hline
\multicolumn{1}{|c|}{Neutrophils}        & \multicolumn{1}{c|}{0.9968}   & \multicolumn{1}{c|}{0.9959}    & 0.9978 & \multicolumn{1}{c|}{0.9421}   & \multicolumn{1}{c|}{0.9092}    & \multicolumn{1}{c|}{0.9775} \\ \hline
\multicolumn{1}{|c|}{Debris/Doublets}    & \multicolumn{1}{c|}{0.9692}   & \multicolumn{1}{c|}{0.9818}    & 0.9570 & \multicolumn{1}{c|}{0.7704}   & \multicolumn{1}{c|}{0.9111}    & \multicolumn{1}{c|}{0.6673} \\ \hline
\multicolumn{1}{|c|}{Monocytes}          & \multicolumn{1}{c|}{0.9679}   & \multicolumn{1}{c|}{0.9571}    & 0.9791 & \multicolumn{1}{c|}{0.8419}   & \multicolumn{1}{c|}{0.8475}    & \multicolumn{1}{c|}{0.8364} \\ \hline
\multicolumn{1}{|c|}{Mature Sig Lambda}  & \multicolumn{1}{c|}{0.9897}   & \multicolumn{1}{c|}{0.9939}    & 0.9856 & \multicolumn{1}{c|}{0.9561}   & \multicolumn{1}{c|}{0.9864}    & \multicolumn{1}{c|}{0.9277} \\ \hline
\multicolumn{1}{|c|}{Mature Sig Kappa}   & \multicolumn{1}{c|}{0.9923}   & \multicolumn{1}{c|}{0.9866}    & 0.9981 & \multicolumn{1}{c|}{0.9421}   & \multicolumn{1}{c|}{0.9421}    & \multicolumn{1}{c|}{0.9421} \\ \hline
\multicolumn{1}{|c|}{TCRgd+}             & \multicolumn{1}{c|}{0.9810}   & \multicolumn{1}{c|}{0.9777}    & 0.9843 & \multicolumn{1}{c|}{0.5549}   & \multicolumn{1}{c|}{0.5698}    & \multicolumn{1}{c|}{0.5408} \\ \hline
\multicolumn{1}{|c|}{TCRgd-}             & \multicolumn{1}{c|}{0.9403}   & \multicolumn{1}{c|}{0.9145}    & 0.9677 & \multicolumn{1}{c|}{0.8634}   & \multicolumn{1}{c|}{0.8195}    & \multicolumn{1}{c|}{0.9122} \\ \hline
\multicolumn{1}{|c|}{CD4+CD8dim}         & \multicolumn{1}{c|}{0.9452}   & \multicolumn{1}{c|}{0.9504}    & 0.9401 & \multicolumn{1}{c|}{0.5899}   & \multicolumn{1}{c|}{0.7111}    & \multicolumn{1}{c|}{0.5039} \\ \hline
\multicolumn{1}{|c|}{NK Cells}           &                               &                                &        & \multicolumn{1}{c|}{0.9313}   & \multicolumn{1}{c|}{0.9433}    & \multicolumn{1}{c|}{0.9196} \\ \cline{1-1} \cline{5-7} 
\multicolumn{1}{|c|}{Myeloid Cells}      &                               &                                &        & \multicolumn{1}{c|}{0.8321}   & \multicolumn{1}{c|}{0.9489}    & \multicolumn{1}{c|}{0.7409} \\ \hline
\multicolumn{1}{|c|}{CD56dim}            & \multicolumn{1}{c|}{0.9827}   & \multicolumn{1}{c|}{0.9683}    & 0.9975 &                               &                                &                             \\ \cline{1-4}
\multicolumn{1}{|c|}{Eosinophils}        & \multicolumn{1}{c|}{0.9722}   & \multicolumn{1}{c|}{0.9713}    & 0.9732 &                               &                                &                             \\ \cline{1-4}
\multicolumn{1}{|c|}{Monocytoid DC}      & \multicolumn{1}{c|}{0.9563}   & \multicolumn{1}{c|}{0.9713}    & 0.9417 &                               & \multicolumn{1}{l}{}           & \multicolumn{1}{l}{}        \\ \cline{1-4}
\multicolumn{1}{|c|}{Basophils}          & \multicolumn{1}{c|}{0.9123}   & \multicolumn{1}{c|}{0.9877}    & 0.8476 &                               & \multicolumn{1}{l}{}           & \multicolumn{1}{l}{}        \\ \cline{1-4}
\multicolumn{1}{|c|}{Neutrophils (U.S.)} & \multicolumn{1}{c|}{0.7632}   & \multicolumn{1}{c|}{0.7106}    & 0.8242 &                               & \multicolumn{1}{l}{}           & \multicolumn{1}{l}{}        \\ \cline{1-4}
\multicolumn{1}{|c|}{Myeloid DC}         & \multicolumn{1}{c|}{0.8908}   & \multicolumn{1}{c|}{0.8413}    & 0.9464 &                               & \multicolumn{1}{l}{}           & \multicolumn{1}{l}{}        \\ \cline{1-4}
\multicolumn{1}{|c|}{CD56bright}         & \multicolumn{1}{c|}{0.8785}   & \multicolumn{1}{c|}{0.9860}    & 0.7921 &                               & \multicolumn{1}{l}{}           & \multicolumn{1}{l}{}        \\ \cline{1-4}
\multicolumn{1}{|c|}{Plasmocytoid DC}    & \multicolumn{1}{c|}{0.7790}   & \multicolumn{1}{c|}{0.8011}    & 0.7581 &                               &                                &                             \\ \cline{1-4}
\multicolumn{1}{|c|}{Plasma Cells}       & \multicolumn{1}{c|}{0.9677}   & \multicolumn{1}{c|}{1.0000}    & 0.9375 &                               &                                &                             \\ \cline{1-4}
\end{tabular}
\end{small}
\end{table*}

\section*{Discussion}
\newpage
From the results shown in \nameref{template_obtention}, particularly  in Figure \ref{clust_templates}, what we see is that optimalFlowTemplates produces trees more similar to the ground truth than the ones given by flowMatch. In particular, we see that optimalFlowTemplates gives a similar structure to the one obtained with $\tilde{d}_{learning}$. Furthermore, optimalFlowTemplates seems to capture the difference between healthy and sick individuals accurately. This comes from the fact that clusters of sick individuals merge other clusters high in the tree. However, we see that this is not entirely the case for flowMatch.  Hence, we can suggest that optimalFlowTemplates captures more accurately the similarity between cytometries than flowMatch. Three additional facts should be stated: first, the similarity distance is independent of parameters, something that is not the case for the generalized edge cover distance used in flowMatch. Second, optimalFlowTemplates produces templates only at one stage, once the number of clusters is determined, while flowMatch produces templates at every stage of the hierarchical clustering procedure. Third, optimalFlowTemplates uses similarity distance which is bound between 0 and 1 and has a clear meaning; however, no bounds are available for flowMatch.

We also notice that optimalFlowTemplates is able to capture differences in the group of healthy and in the group of sick individuals. Therefore, it seems to be capturing additional biological and/or technical variability and not only the one provided by the distinction between healthy and sick. This can be clearly seen in Table \ref{database_clustering}, where we have several clusters for healthy individuals and also several clusters for sick individuals. Hence, we can infer that similarity distance is sensitive enough to differentiate between cytometries and therefore that it is a suitable distance for comparing them.

With respect to the results shown in section \nameref{results_gating}, there are several interesting implications. Firstly, as expected, our supervised method, optimalFlowTemplates + optimalFlowClassification outperforms an unsupervised method as flowMeans, as seen from the higher values in each entry of column three, with respect to column one, in Table \ref{comp_classif}. We also see that the F-measures obtained by our procedure are very satisfactory giving a mean value of 0.9539 and a median value of 0.9587 for $\mathcal{TS}$. It is also worth noting that good results are obtained for healthy individuals and for the patients with cancer.

Secondly, a comparison with a supervised method, deepCyTOF, based on neural networks and domain adaptation has been provided. We want to stress that, at least with the implementation provided in github, we were unable to apply deepCyTOF to cytometries with different cell types. Hence we had to produce a modified test group given by $\mathcal{TS}'$. This favours deepCyTOF since our procedure is applied to the original cytometries. This limited its applicability since it did not accommodate cytometries with different cell types.  From column two, in Table \ref{comp_classif}, we see that deppCyTOF works well for cytometries $\{\mathcal{C}^{2'},\mathcal{C}^{5'},\mathcal{C}^{14'},\mathcal{C}^{15'}\}$ giving results comparable to our own procedure. However, it does not work as well for cytometries $\{\mathcal{C}^{7'}, \mathcal{C}^{17'}, \mathcal{C}^{18'}\}$. We see that our procedure is better for every cytometry in $\mathcal{TS}'$, even in the previously mentioned disadvantageous position, except for $\mathcal{C}^{14'}$. We stress that deepCyTOF is using precisely $\mathcal{C}^{14'}$ for learning, as indicated in italic in column 2 of Table \ref{comp_classif}, and therefore it is using 80\% of the data in $\mathcal{C}^{14'}$ to calibrate the neural-network. Hence it is natural that deepCyTOF is the best procedure for this cytometry.

Thirdly, as a consequence of the versatility of our procedure optimalFlowTemplates, we can use it to improve the results of deepCyTOF. In essence, deepCyTOF's domain adaptation is not able to account for the high variability in $\mathcal{TS}'$, but this is exactly what optimalFlowTemplate is for. We start by creating a set of gated cytometries by gating in an unsupervised fashion, using flowMeans, the cytometries in $\mathcal{TS}'$. Then we apply optimalFlowTemplates and obtain the tree shown in Figure \ref{clust_deepcytof}. This suggests spliting the original group of cytometries into three different clusters $\mathcal{TS}'_1$, $\mathcal{TS}'_2$ and $\mathcal{TS}'_3$. Now, applying deepCyTOF to each cluster separately, we obtain the results shown in column three of Table \ref{comp_classif}. What we see is a significant improvement in the results of deepCyTOF for cytometries $\{\mathcal{C}^{7'},\mathcal{C}^{17'},\mathcal{C}^{18'}\}$ and comparable results for  $\{\mathcal{C}^{2'}, \mathcal{C}^{5'},\mathcal{C}^{14'},\mathcal{C}^{15'}\}$. Hence, our procedure has indeed helped to improve the performance of deepCyTOF.

It is important to note that when deepCyTOF learns in a cytometry, as is the case for $\mathcal{C}^{14'}$, and $\{\mathcal{C}^{2'},\mathcal{C}^{7'},\mathcal{C}^{17'}\}$, for columns two and three of Table \ref{comp_classif}, respectively, it uses 80\% of the sample. This justifies the great result for $\mathcal{C}^2$ and $\mathcal{C}^7$ in column three of Table \ref{comp_classif}.

Additionally, we want to highlight something that is reflected in Table \ref{table_medain_f}. Both cytometries, taken from individuals with cancer, $\{\mathcal{C}^{26},\mathcal{C}^{27}\}$ are assigned to clusters of cytometries of patients with cancer. Therefore, our procedure is correctly assigning sick individuals to clusters of sick individuals. Furthermore, when in the cluster there is a patient with the same type of cancer results are very good. This is the case for $\mathcal{C}^{27}$ where it is assigned to cluster 5 in the third column of Table \ref{database_clustering}, where $\mathcal{C}^{28}$ also has Chronic Lymphocytic Leukemia. It is also worth noting that healthy individuals are assigned to clusters of healthy individuals.

The last thing we want to discuss is the meaning of the median F-measure values, as the ones given in table \ref{table_medain_f}, which are obtained from values such as the ones shown in Table \ref{f_measure_cell_types}. What we see is that a high median F-measure value indicates that classification is good along all cell types, regardless of the number of cells in each cell type. This is quite important since often cell types with a small number of cells are very relevant for diagnosis. Indeed, from the values in Table \ref{f_measure_cell_types} we see that our procedure is giving a good performance in almost all cell types.

\section*{Conclusion}
Efficiency, robustness, scalability and accountability are fundamental requisites for a viable automated supervised gating workflow. We think that we have provided such a method using as a baseline optimal-transport-techniques.

Efficiency is ensured through the use of automatically produced clusters of previously gated cytometries that capture well information as sickness and other type of variability. The grouping of cytometries and the automated assignation of a new cytometry to the most similar group also accounts for a robust procedure.

Since manual gating is avoided or may be required just in a few instances, our procedures can be applied to big data sets with only computation cost as a burden.

Accountability emerges from the way that we have approached the problem. Cluster memberships can be screened and, furthermore, synthetic prototypes can be manually gated in order to check their suitability. Hence when we classify, even with some black-box procedure, we have an accurate understanding of the starting point.

In our view, an additional benefit of our procedure is versatility. It can incorporate many previously existing tools in automated flow cytometry gating and it can accommodate many supervised learning procedures.

\section*{Abbreviations}
\textbf{BM:} Bone Marrow\\
\textbf{CLL:} Chronic Lymphocytic Leukimia\\
\textbf{DLBCL:} Diffuse Large B-cell Lymphoma\\
\textbf{FC:} Flow Cytometry\\
\textbf{FL:} Follicular Lymphoma\\
\textbf{HCL:} Hairy Cell Leukimia\\
\textbf{HD:} Healty Diagnosis\\
\textbf{LN:} Lymph Node\\
\textbf{LPL:} Lymphoplasmacytic Lymphoma\\
\textbf{MCL:} Mantle Cell Lymphoma\\
\textbf{PB:} Peripherial Blood\\
\textbf{QDA:} Quadratic Discriminant Analysis\\

%\newpage

%%%%%%%%%%%%%%%%%%%%%%%%%%%%%%%%%%%%%%%%%%%%%%
%%                                          %%
%% Backmatter begins here                   %%
%%                                          %%
%%%%%%%%%%%%%%%%%%%%%%%%%%%%%%%%%%%%%%%%%%%%%%

\section*{Ethics approval and consent to participate}
Data used in this work were obtained under the project with acronym EuroFlow which was approved by the ethics comity \emph{Comité de Ética Asistencial del Área de Salud de Salamanca (Hospital Universitario de Salamanca, Salamanca (Spain))} in an ordinary session on February the 26th of 2007. The participants granted permission for their data to be used for research purposes.
\section*{Consent for publication}
Not applicable.
\section*{Availability of data and materials}
With the user in mind we have developed an R package called optimalFlow that implements our methodology which is available in the Bioconductor repository at\\
\href{https://bioconductor.org/packages/optimalFlow}{https://bioconductor.org/packages/optimalFlow}. The data used in this work is freely available for researchers after registration on \href{https://www.EuroFlow.org}{https://www.EuroFlow.org}.
\section*{Competing interests}
The authors declare that they have no competing interests.
\section*{Funding}
Research partially supported by FEDER, Spanish Ministerio de Economía y Competitividad, grant MTM2017-86061-C2-1-P and Junta de Castilla y León, grants VA005P17 and VA002G18. Research partially supported by DEEL IRT. We thank the AI interdisciplinary institute ANITI, grant agreement number ANR-19-PI3A-0004 under the French investing for the future PIA3 program. The funding bodies did not play any role in the design of the methodology, creation of the algorithms, analysis and interpretation of data, or in writing the manuscript. 
\section*{Author's contributions}
All authors conceived and designed the study. HI designed the algorithms and wrote the
manuscript. All authors read and approved the final manuscript.
\section*{Acknowledgements}
The motivation of this work is due to a collaboration with the clinical research team, led by Dr. Alberto Orfao, of the Cancer Research Center in Salamanca. We are extremely grateful for their insights and help, as well as for all the data they have kindly allowed us to use in this work.

\appendix

\section*{Notions on optimal transport}\label{optimal_transport}
Following \cite{opt-trans}, let us take $\mathcal{P}(\Omega)$, the space of probability distributions on $\Omega$. For $\mu,\nu$ in $\mathcal{P}(\Omega)$, let us define $\Pi(\mu,\nu)$ the set of all probability measures $\pi$ on $\Omega\times\Omega$ with first marginal $\mu$ and second marginal $\nu$. The optimal transport cost between the two measures is defined as
\begin{equation}
\label{ot cost}
C(\mu,\nu) = \inf_{\pi\in\Pi(\mu,\nu)}\int c(x,y)d\pi(x,y)
\end{equation} 
where $c(x,y)$ is the cost of transporting one unit of mass from $x$ to $y$. A probability $\pi$ that achieves the minimum in (\ref{ot cost}) is called an optimal coupling, with an associated random variable $(X,Y)$ that has joint distribution $\pi$. When $\mu$ and $\nu$ are discrete, i.e., $\mu=\sum_{i=1}^np_i\delta_{x_i}$ and $\nu=\sum_{j=1}^mq_i\delta_{y_i}$, with $x_i,y_j\in\mathbb{R}^d$, the optimal transport problem can be solved as a linear program (see \cite{opt-trans_2}) where
\begin{equation*}
C(\mu,\nu)=\sum_{i=1}^{n}\sum_{j=1}^{m}w^*_{ij}c(x_i,y_j),
\end{equation*}
and $(w^*_{ij})$ are the solutions of the optimal transport linear program 
\begin{small}
\[\begin{array}{ll@{}ll}
\text{minimize} & \sum_{i=1}^{n}\sum_{j=1}^{m}w_{ij}c(x_i,y_j)&\\
\text{subject to} & w_{ij}\geq 0,  &1\leq i\leq n,1\leq j\leq m&\\ 
				  & \sum_{j=1}^{m}w_{ij} = p_i, & 1\leq i\leq n&\\
				  & \sum_{i = 1}^{n}w_{ij} = q_j, & 1\leq j\leq m&\\
				  & \sum_{i=1}^{n}\sum_{j=1}^{m}w_{ij} = 1.
\end{array}\]
\end{small}

For $(\Omega=\mathbb{R}^d, \|\cdot\|)$, with $\|\cdot\|$ the Euclidean norm, and $p\in [1,\infty)$, the $p-$Wasserstein distance between $\mu$ and $\nu$ is defined as
\begin{small}
\begin{align*}
\mathcal{W}_p^p(\mu,\nu) &= \inf_{\pi\in\Pi(\mu,\nu)}\int\|x-y\|^p d\pi(x,y)\\
&=\inf\left\{E\|X-Y\|^p, \mathcal{L}(X)=\mu, \mathcal{L}(Y) = \nu\right\},
\end{align*}
\end{small}
where $\mathcal{L}(X)$ refers to the law of $X$.

We present the entropy regularized Wasserstein distance, since it is strictly convex and there are efficient solutions based on the Sinkhorn algorithm (see \cite{cuturi2014}). For a fixed $\gamma>0$ the regularized Wasserstein distance is defined as
\begin{equation}\label{entropy_wasserstein}
\mathcal{W}_\gamma(\mu,\nu)=\sum_{i=1}^{n}\sum_{j=1}^{m}w^*_{ij}\|x_i-y_j\|^2 + \gamma\sum_{i=1}^n\sum_{j=1}^mw^*_{ij}\log w^*_{ij},
\end{equation}
where $(w^*_{ij})$ are the solutions of the optimal transport linear program 

\[\begin{array}{ll@{}ll}
\text{minimize} & \sum_{i=1}^{n}\sum_{j=1}^{m}w_{ij}\|x_i-y_j\|^2 + \gamma\sum_{i=1}^{n}\sum_{j=1}^{m}w_{ij}\log w_{ij}&\\
\text{subject to} & w_{ij}\geq 0, &  1\leq i\leq n,1\leq j\leq m&\\ 
				  & \sum_{j=1}^{m}w_{ij} = p_i, & 1\leq i\leq n&\\
				  & \sum_{i = 1}^{n}w_{ij} = q_j, & 1\leq j\leq m&\\
				  & \sum_{i=1}^{n}\sum_{j=1}^{m}w_{ij} = 1.
\end{array}\]

Let us denote $\mathcal{P}_2(\mathbb{R}^d)$ the set of probability measures on $\mathbb{R}^d$ with finite second moment and let us consider $\mathcal{W}_2(\mu,\nu)$ for $\mu,\nu \in \mathcal{P}(\mathbb{R}^d)$. In \cite{bary_wassesr_rob} the notions of $k$-barycenter and trimmed $k$-barycenter were introduced, building on the concept of Wasserstein barycenter introduced in \cite{boissard2015, bary_wasser}. A $k$-barycenter of probabilities $\{\mu_1,\dots,\mu_n\}$ in $\mathcal{P}_2(\mathbb{R}^d)$ with weights $\lambda_1,\dots,\lambda_n$ is any k-set $\{\bar{\mu}_1,\dots,\bar{\mu}_k\}$ in $\mathcal{P}_2(\mathbb{R}^d)$ such that for any $\{\nu_i,\dots,\nu_k\}\subset \mathcal{P}_2(\mathbb{R}^d)$ we have that
\begin{small}
\begin{equation}
\label{k_bar_1}
\sum_{i = 1}^n \lambda_i\min_{j\in\{1,\dots,k\}}\mathcal{W}^2_2(\mu_i,\bar{\mu}_j)\leq \sum_{i = 1}^n \lambda_i\min_{j\in\{1,\dots,k\}}\mathcal{W}^2_2(\mu_i,\nu_j).
\end{equation}
\end{small}
An $\alpha$-trimmed $k$-barycenter of $\{\mu_1,\dots,\mu_n\}$ with weights as before is any  k-set $\{\bar{\mu}_1,\dots,\bar{\mu}_k\}$ with weights $\bar{\lambda} 
=(\bar{\lambda}_1,\dots,\bar{\lambda}_n)\in \Lambda_\alpha(\lambda)$ such that
\begin{equation}\label{trimmed_k_bar_1}
\sum_{i = 1}^n \bar{\lambda}_i\min_{j\in\{1,\dots,k\}}\mathcal{W}^2_2(\mu_i,\bar{\mu}_j)= \min_{\{\nu_1,\dots,\nu_k\}\subset \mathcal{P}_2(\mathbb{R}^d), \lambda^*\in \Lambda_\alpha(\lambda)}\sum_{i = 1}^n \lambda^*_i\min_{j\in\{1,\dots,k\}}\mathcal{W}^2_2(\mu_i,\nu_j),
\end{equation}
where $\Lambda_\alpha(\lambda)=\{\lambda^* = (\lambda^*_1,\dots,\lambda^*_n):$ $0\leq \lambda^*_i\leq \lambda_i/(1-\alpha), \sum_{i=1}^n\lambda^*_i = 1\}$.

Broadly speaking k-barycenters can be thought of as an extension of k-means to the space of probabilities with finite second order, since we can rewrite (\ref{k_bar_1}) as
\begin{equation}\label{bar_k-means}
\min_{\mathfrak{S}}\sum_{j=1}^k\sum_{\mu_i\in\mathfrak{S}_j}\lambda_i \mathcal{W}^2_2(\mu_i,\bar{\mu}_j)
\end{equation}
where $\mathfrak{S} = \left\{\mathfrak{S}_1,\dots,\mathfrak{S}_k\right\}$ is a partition of $\{\mu_1,\dots,\mu_n\}$ and $\bar{\mu}_j$ is the barycenter of the elements in $\mathfrak{S}_j$. Therefore, trimmed k-barycenters may be matched to trimmed k-means. As stated in \cite{bary_wassesr_rob}, efficient computations can be done when dealing with location-scatter families of absolutely continuous distributions in  $\mathcal{P}_2(\mathbb{R}^d)$. A notable example being the family of multivariate Gaussian distributions.

%%  After submission of the .TEX file,                     %%
%%  you will be prompted to submit your .BBL file.         %%

% if your bibliography is in bibtex format, use those commands:
\bibliographystyle{plain} % Style BST file (bmc-mathphys, vancouver, spbasic).
\bibliography{thesis}     % Bibliography file (usually '*.bib' )
\end{document}